\pdfoutput=1

\documentclass[11pt]{article}

\usepackage[final]{acl}

\usepackage{times}
\usepackage[most]{tcolorbox}
\usepackage{latexsym}
\usepackage{multirow}
\usepackage{amsfonts} 
\usepackage{amsmath}
\usepackage{tcolorbox}
\tcbuselibrary{skins}

\usepackage[T1]{fontenc}

\usepackage[utf8]{inputenc}

\usepackage{microtype}

\usepackage{inconsolata}
\usepackage{times}
\usepackage{latexsym}
\usepackage{booktabs}
\usepackage{amsmath}
\usepackage{url}
\usepackage{graphicx}

\newcommand{\tabincell}[2]{\begin{tabular}{@{}#1@{}}#2\end{tabular}} 
\DeclareSymbolFont{extraup}{U}{zavm}{m}{n}
\DeclareMathSymbol{\varheart}{\mathalpha}{extraup}{86}
\DeclareMathSymbol{\vardiamond}{\mathalpha}{extraup}{87}
\title{
\textsc{WebCoT}: Enhancing Web Agent Reasoning by Reconstructing Chain-of-Thought in Reflection, Branching, and Rollback
}

\author{Minda Hu$^{\clubsuit\spadesuit}$\thanks{Equal Contribution}
, Tianqing Fang$^{\spadesuit*}$, Jianshu Zhang$^\varheart$, Junyu Ma$^\spadesuit$, Zhisong Zhang$^\spadesuit$, \\
\textbf{Jingyan Zhou$^\clubsuit$, Hongming Zhang$^\spadesuit$, Haitao Mi$^\spadesuit$, Dong Yu$^\spadesuit$, Irwin King$^\clubsuit$}\\
$^\clubsuit$Chinese University of Hong Kong, $^\spadesuit$Tencent AI Lab, $^\varheart$Wuhan University\\
\texttt{\{mindahu21, king\}@cse.cuhk.edu.hk}, \texttt{tianqfang@tencent.com}}

\begin{document}
\maketitle
\begin{abstract}
Web agents powered by Large Language Models (LLMs) show promise for next-generation AI, but their limited reasoning in uncertain, dynamic web environments hinders robust deployment.
In this paper, we identify key reasoning skills essential for effective web agents, i.e., \textit{reflection \& lookahead}, \textit{branching}, and \textit{rollback}, and curate trajectory data that exemplifies these abilities by reconstructing the agent's (inference-time) reasoning algorithms into chain-of-thought rationales.
We conduct experiments in the agent self-improving benchmark, OpenWebVoyager, and demonstrate that distilling salient reasoning patterns into the backbone LLM via simple fine-tuning can substantially enhance its performance. 
Our approach yields significant improvements across multiple benchmarks, including WebVoyager, Mind2web-live, and SimpleQA (web search), highlighting the potential of targeted reasoning skill enhancement for web agents.
\end{abstract}

\section{Introduction}
The rise of large language models (LLMs) has sparked significant interest in developing intelligent agents capable of interacting with the web through a browser, commonly referred to as web agents~\cite{yaoreact, manus, openmanus2025}. 
However, despite recent advancements, even the best-performing web agents still lag far behind human performance—even when compared to users unfamiliar with a website's structure or functionality~\cite{MMInA, GAIA, song2025bearcubs}.
This performance gap is primarily attributed to the limited reasoning abilities of current language models when applied to web agent workflows.

\begin{figure}[t]
    \centering
    \includegraphics[width=0.48\textwidth]{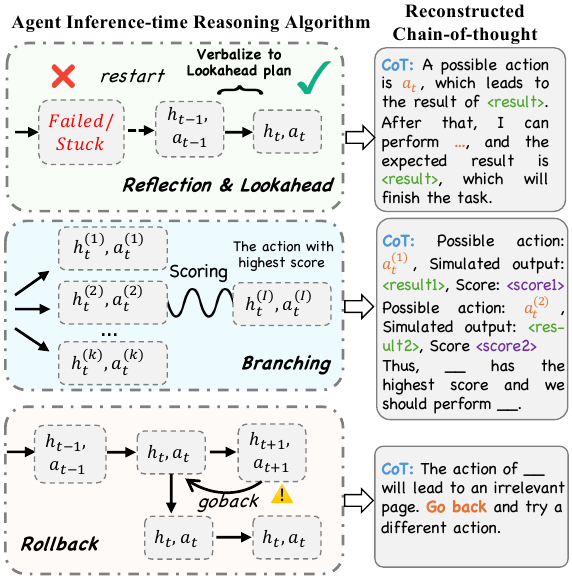} 
    \caption{Overview of our framework. We leverage a language model to translate inference-time processes, i.e., \textit{reflection and look-ahead}, \textit{branching}, and \textit{rollback}, into natural language chain-of-thoughts, which are then used to train the agent language model. }
    \label{fig:overview}
\end{figure}

Despite recent advances in Large Reasoning Models (LRM, e.g., DeepSeek-R1; \citealp{deepseekr1}, QwQ; \citealp{qwen2025qwq32b}), these models primarily focus on arithmetic reasoning and
are prone to overthinking and generating unnecessarily complex solutions for agent tasks~\cite{DBLP:journals/corr/abs-2502-08235,kumar2025overthinking,su2025between}.
While directly applying Reinforcement Learning (RL) in agentic environments~\cite{webrl, webthinker, liu2025infigui, singh2025agentic, wei2025webagentr1trainingwebagents} is a viable alternative, the resulting reasoning abilities are often unpredictable and lack structured priors. 
Moreover, these approaches incur prohibitively high costs~\cite{xu2025towards,dang2025reinforcement} when conducting real-world rollouts. Additionally, they often focus on static and deterministic environments such as WebArena~\cite{webarena}, whereas applying them to stochastic real-world open-domain web environments can be problematic due to the randomness inherent in rollouts.
In contrast, distilling specific reasoning patterns into agents~\cite{chen2024magdi,zhao2024we,hu2025ahasystematicmetaabilitiesalignment} combines the adaptability of learned policies with the interpretability and task-aware heuristics of curated reasoning, mitigating both the overthinking problem and the exploration burden of pure RL.

In this paper, we carefully examine and design the specific reasoning abilities required for effective web agents, and sample corresponding agent trajectories to conduct Supervised Fine-Tuning (SFT) on LLMs.
In particular, we focus on three key components: (1) \textit{reflection \& lookahead}, the ability to reflect on previous failures and conjure precise long-horizon plans. 
(2) \textit{branching}, the ability to sample multiple possible actions and have accurate awareness of the possible outcomes, selecting the most promising one; 
and (3) \textit{rollback}, the ability to validate error, roll back to a previous state, and self-correct the agent's mistake. 
To study these abilities, we implement representative algorithms for each component: a novel reflection-and-lookahead module, WebDreamer~\cite{webdreamer} for \textit{branching}, and AgentRollback~\cite{zhang2025enhancingwebagentsexplicit} for \textit{rollback}.
Simple illustrations of the three abilities are shown in Figure~\ref{fig:overview}.
For \textit{reflection and look-ahead}, 
we analyze failed trajectories or redundant steps, distill the corrected trajectories after reflecting on errors, into key planning steps, and further refine them into structured chain-of-thought rationales using an LLM.
For \textit{branching}, we sample multiple actions and select the best one using an LLM, and then paraphrase this selection process into a paragraph of rationale.
For \textit{rollback}, we reference a successful trajectory and sample an intermediate state to intentionally generate incorrect branches, thereby constructing a \textit{goback} action along with its corresponding rationale.

To validate the effectiveness of the proposed method, experiments are conducted following the settings in the web agent self-improvement benchmark, OpenWebVoyager~\cite{openwebvoyager}. 
Specifically, we use \textsc{Llama-3.3}~(\citealp{llama3}; 70B) as the backbone LLM for sampling web agent trajectories, and employ GPT-4o to perform rationale paraphrasing and action selection in \textit{reflection \& look-ahead}, \textit{branching}, and \textit{rollback}. 
The collected trajectories are then used to fine-tune \textsc{Llama-3.3} as our final agent model.
Our results demonstrate that distilling these reasoning patterns into LLM chain-of-thought yields significant performance gains across multiple benchmarks, such as WebVoyager~\cite{webvoyager}, Mind2Web-Live~\cite{webcanvas}, and SimpleQA~\cite{simpleqa}.
The approach substantially outperforms rejection sampling baselines and even exceeds distillation from more capable LRM like QwQ-32B~\cite{qwen2025qwq32b}. 
Overall, our research\footnote{Information related to WebCoT dataset and code implementations can be found in \url{https://github.com/Tencent/SelfEvolvingAgent}.} underscores the importance of explicitly defining and instilling effective thinking and reasoning patterns for agent tasks.

\section{Related Works}
\paragraph{Web Agents.}
Leveraging advanced backbone LLMs~\citep{llama3, leopard, gpt4, claude37}, web agents demonstrate proficiency in interacting effectively with diverse web environments, solving various tasks within specialized frameworks~\citep{yaoreact, cognitive_kernel, fang2025cognitive}. Recent research increasingly explore data-driven methodologies~\citep{Explorer, sun2025osgenesisautomatingguiagent, xu2024agenttrek, DBLP:journals/corr/abs-2502-06776} to further enhance the performance of open-source models in web-based tasks. Additionally, several studies focus on equipping agents with self-improvement mechanisms~\citep{fang2025webevolver, rest_meets_react, LLM_agent_can_self_improve, zhang2025enhancingwebagentsexplicit}, enabling models to iteratively refine their strategies through bootstrapped learning. To more comprehensively evaluate and promote advancements in web agent capabilities, numerous benchmarks are proposed to assessing performance across diverse web-related tasks~\citep{webshop, webarena, webvoyager, WebWalker, MMInA}.

\paragraph{Agents with Reasoning.}
Integrating advanced reasoning mechanisms into agents has garnered substantial attention, significantly enhancing their performance. Earlier frameworks~, such as ReAct~\citep{yaoreact}, SeRTS~\citep{hu-etal-2024-serts}, and Reflexion~\citep{Reflexion}, introduce iterative reasoning-action loops to improve agent decision-making. Leveraging extended test-time scaling (TTS), recent models like OpenAI-o1~\citep{openai2024openaio1card}, Qwen-QwQ~\citep{qwen2025qwq32b}, and DeepSeek-R1~\citep{deepseekr1} have demonstrated remarkable performance improvements through explicit chain-of-thought reasoning~\citep{wei2022chain}. Several studies, including WebThinker~\citep{webthinker}, LAMs~\citep{zhang2025agent}, and Agent-Reasoning~\citep{wu2025agentic}, have further showcased that TTS can elevate the performance ceiling for agents. 
Similar to the observations in Meta Ability~\cite{hu2025ahasystematicmetaabilitiesalignment} that RL is difficult to control and ``aha'' behaviors remain unpredictable, both TTS and Reinforcement Learning alone struggle to equip agents with our targeted capabilities such as reflection, branching, and rollback.
To address this limitation, we propose \textsc{WebCoT}, a method that leverages carefully curated chains of thought exemplifying essential reasoning skills, thereby facilitating improved reasoning ability in web agents.

\section{Preliminary}

This section formalizes the web agent task and presents the foundational components of our agent optimization framework.
\subsection{Problem Formulation}
The web agent task is modeled as a Partially Observable Markov Decision Process (POMDP), defined by $(S, \mathcal{A}, \mathcal{O}, \mathcal{T}, \mathcal{R})$. 
{State (\(S\))} represents the state of the web environment, with \(s_t\) at step \(t\). {Action (\(\mathcal{A}\))} includes atomic web operations such as \texttt{click}, \texttt{type}, \texttt{goback}, \texttt{scroll}, and \texttt{stop} \citep{webvoyager}. {Observation (\(\mathcal{O}\))} captures visible elements of the environment, with \(o_t = \Omega(s_t)\), where \(\Omega\) extracts content like accessibility trees. {Transition (\(\mathcal{T}\))} advances the state \(s_t\) deterministically based on browser operations. {Reward (\(\mathcal{R}\))} provides evaluations on the agent trajectories.

The agent processes a natural language query \(q\) requiring multi-step interactions in a web environment.
The agent's policy \(\pi(o_t, q) \to a_t\) generates actions \(a_t\) based on the query \(q\) and the current observation \(o_t\), forming a trajectory \(\tau = \{(o_1, a_1), \ldots, (o_t, a_t)\}\). Rewards are computed using a self-assessment function \(\hat{r}(\tau, q) \in [0, 1]\).

For web navigation, given a query \(q\) and target website \(w\), the environment is initialized, and the first observation \(o_1\) is obtained. Following \textit{Cognitive Kernel}~\cite{cognitive_kernel}, the accessibility tree represents \(o_t\). A Large Language Model (LLM), parameterized by \(\theta\), serves as the policy network, generating Chain-of-Thought reasoning \(h_t\) and actions \(a_t\):
\begin{equation}
    (h_t, a_t) \sim \pi_\theta(\cdot \mid I, q, o_{1:t}, h_{1:t-1}, a_{1:t-1}),
    \tag{1}
\end{equation}
where \(I\) denotes system instructions. The environment evolves based on:
\begin{equation}
    s_{t+1} = \mathcal{T}(s_t, a_t), \quad o_{t+1} = \Omega(s_{t+1}),
    \tag{2}
\end{equation}
producing a trajectory \(\tau = \{(o_i, h_i, a_i)\}_{i=1}^T\), where \(T\) is the total number of steps.

\subsection{Optimization}
We adopt a self-improvement optimization framework as in OpenWebVoyager~\cite{openwebvoyager}.
We introduce the backbone agent foundation model, denoted as $\mathcal{M}$, along with its corresponding policy function, $\pi_{\mathcal{M}}$. The model $\mathcal{M}$ is used to sample actions based on a given input query $q$, which are subsequently utilized to collect web navigation trajectories. As the core of the \textit{Cognitive Kernel}, $\mathcal{M}$ enables interactions with the web environment. To inform its decisions, the agent observes the past $m$ steps of interaction, represented as webpage accessibility trees.

For each query \( q \in \mathcal{Q} \), the set of all queries, a trajectory \(\tau_i\) is sampled from the policy \(\pi_{\theta_{\mathcal{M}}}(\tau \mid I, q)\). To mitigate performance degradation caused by excessively long contexts, we clip the trajectory history \( c_t \) when \( t - 1 > k \), retaining only the most recent $k$ observations. Thoughts and actions are preserved, as they contain compressed information about the history:
\begin{equation}
\begin{aligned}
    c_t^{\text{clip}} = \big(&h_1, a_1, h_2, a_2, \ldots, h_{t-k}, a_{t-k}, \\
    &o_{t-k+1}, h_{t-k+1}, a_{t-k+1}, \ldots, o_{t-1} \big),
\end{aligned}
\end{equation}
such that the new actions are generated with the following function:
\begin{equation}
    (h_t, a_t) \sim \pi_{\theta_{\mathcal{M}}}(\cdot \mid I, q, c_t^{\text{clip}}).
\end{equation}
For a train set with collected trajectories $\mathcal{D} = \{(q'_i, \tau'_i)\}^t_{i=1}$, we aim to maximize the following objective function:
\begin{equation}
\begin{aligned}
    \mathcal{J}(\theta) = \mathbb{E}_{(q', \tau') \sim \mathcal{D}} \Bigg[ \sum_{t=1}^{T} & \Big( 
    \log \pi_\theta(a_t \mid q, c_t^{\text{clip}'}, h_t) \nonumber \\
    & + \log \pi_\theta(h_t \mid q, c_t^{\text{clip}'}) 
    \Big) \Bigg],
\end{aligned}
\end{equation}
to refine the training data, a rejection sampling dataset \(\mathcal{D}_{\text{rej}}\) is constructed by filtering and retaining only trajectories that satisfy an automatic evaluation metric \(r(\tau, q)\).

\begin{figure*}[t]
    \centering
    \includegraphics[width=1.0\textwidth]{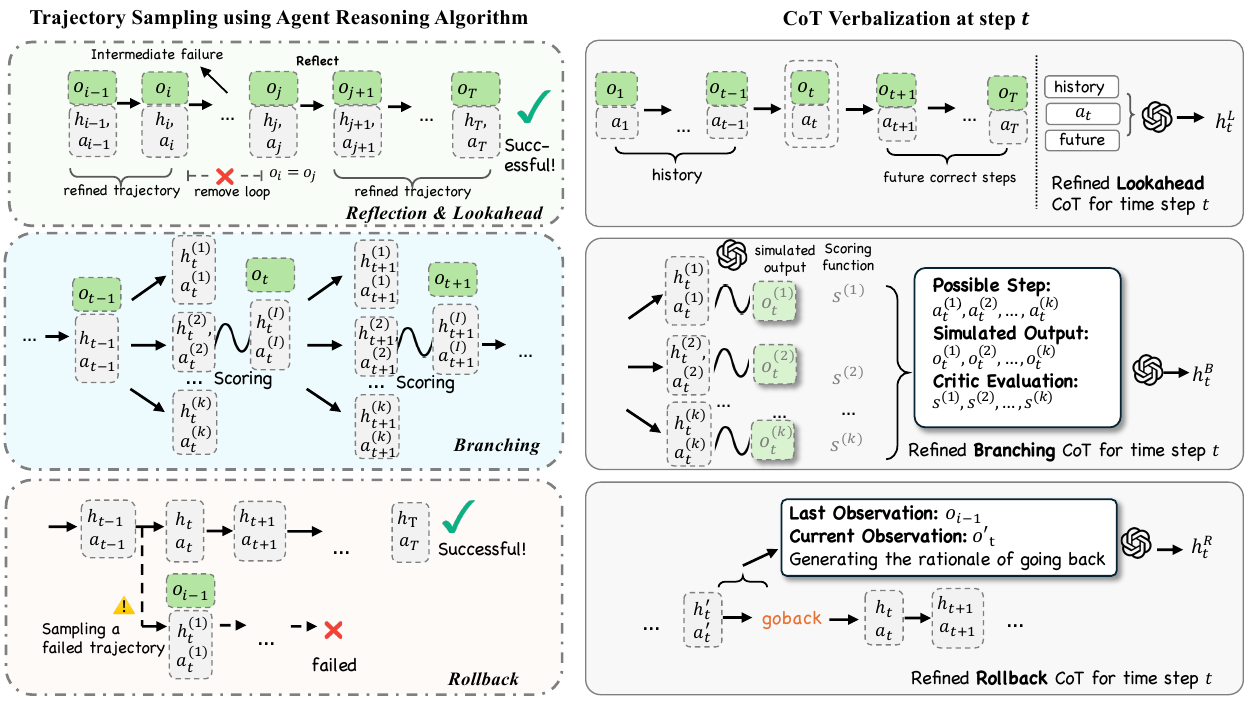} 
    \caption{Overview of the components in \textsc{WebCoT}. In \textit{Reflection \& \textbf{L}ookahead}, intermediate failures are identified and removed. Verbalized lookahead planning rationales ($h_t^L$) are then used as a reconstructed chain-of-thought to perform reflection. In \textit{\textbf{B}ranching}, the process of sampling and scoring alternative actions is verbalized as a chain-of-thought ($h_t^B$). In \textit{\textbf{R}ollback}, trajectories that require returning to a previous successful state are deliberately constructed, and the rationale for rollback is distilled as $h_t^R$.}
    \label{fig:method}
\end{figure*}

\section{\textsc{WebCoT}}
In this section, we introduce the details of the \textsc{WebCoT} reasoning patterns, \textit{reflection \& lookahead}, \textit{branching}, and \textit{rollback}, and how to perform \textit{cumulative training} on top of the OpenWebVoyager self-improvement framework.
An overview of the pipeline is shown in Figure~\ref{fig:method}.

\subsection{Reflection \& Lookahead}
\label{sec:tnf}
We start by identifying trajectories in $\mathcal{D}$ by detecting intermediate failures.
Then, we generate Chain-of-thought such that the agent can reflect from previous mistakes and make better look-ahead planning~\cite{zhang2025streaminglookingaheadtokenlevel}.

We identify intermediate errors by detecting trajectory loops, which emerge when either failed actions or logical inconsistencies produce repeated observations across different time steps.
Formally, we denote such trajectories as $\tau^{loop} = \{(o_t, h_t, a_t)\}_{t=1}^T$, where exists $o_i = o_j$ and $i<j$, 
forming a loop.

First, we refine the trajectory by removing the redundant sequences between $o_i$ and $o_j$.
Then, we re-generate chain-of-thoughts in the refined trajectory to guide the agent LLM toward improved reflection and look-ahead planning.
Specifically, 
we verbalize the refined successful trajectory to an abstract plan using GPT-4o-mini\footnote{We find that GPT-4o-mini is capable enough for the simple job of verbalization} to replace the original chain-of-thoughts $\{h_t\}^T_{t=1}$.
At a current time step $t_c$, a plan is generated by refining the trajectory history $\tau_{t<t_c}$
, current observation $o_{t_c}$, current action $a_{t_c}$, and future trajectory 
$\tau_{t>t_c}$
. The detailed prompt template in Table~\ref{tab:lookahead_cot_prompt} in the Appendix.
The original chain-of-thought $h_{t_c}$ is then replaced with the generated \textbf{L}ookahead planning guidance $h^L_{t_c}$ to enhance the agent's planning ability.
The resulting new trajectory are denoted as $\tau^{L} = \{(o_i, h^{L}_t, a_t)\}^{T-|i-j|+1}_{t=1}$, where $h^{L}_t$ is the newly verbalized chain-of-thought.

We denote the resulting refined trajectories as $\mathcal{D}^L$, indicating that they contain chain-of-thoughts which both reflect on previous errors and incorporate \textbf{L}ookahead planning as part of their rationale.

\subsection{Branching}
\label{sec:sim_branch}
The second reasoning enhancement, branching, focuses on the ability to sample several possible actions and determine the best one leading to success.
Tree search-based planning with real-world interactions often incurs high computational costs and risks irreversible actions, whereas simple Model Predictive Control (MPC) can serve as an efficient and effective substitute (WebDreamer; \citealp{webdreamer}).
We use a similar implementation as in WebDreamer, to simulate possible future states for each action sampled over a finite horizon using a function $sim(o, a)$.
Then, we score the actions with a scoring function $score(sim(o_t, a_t^{(i)}))$, and execute the action $a_I$ with the highest score $I = \mathop{\arg\max}_{i} score(sim(o_t, a_t^{(i)}))$. 
The process repeats after observing new states, allowing the agent to adapt dynamically while minimizing unnecessary interactions.
In detail, at a time step $t$, we
generate $k$ candidate actions $\{(h_t^{(i)}, a_t^{(i)})\}^{k}_{i=1}$ for a given observation $o_t$, simulate two-step future prediction for each action, and select the action with the highest score to execute. 
The scoring function and simulation function are all based on GPT-4o. We leave the detailed prompts to Appendix \ref{sec:webdreamer}. 

Despite its strengths, this approach still incurs significant inference overhead. 
We aim to condense the sophisticated multi-turn prompting pipeline into a single chain-of-thought paragraph, thereby enhancing the agent's branching ability.
We begin by sampling new trajectories on the original set of query using WebDreamer, resulting in trajectories $\tau^{B} = \{(o_t, h^{B}_t, a^{B}_t)\}^T_{t=1}$. 
We then refine the reasoning chain-of-thoughts $\{h^{B}_t\}^T_{t=1}$ by following the template in Table~\ref{tab:branch_prompt}, verbalizing the action sampling and selection process as natural language rationale.
Finally, we collect all $\tau^{B}$ to obtain the set of trajectories that were successfully executed using WebDreamer, representing the ability to perform \textbf{B}ranching. We denote this set as $\mathcal{D}^{B} = \{(q, \tau^{B})\}$.

\subsection{Rollback}\label{sec:backtracking}
Rollback introduces a complementary mechanism to further enhance the reasoning and decision-making capabilities of $\mathcal{M}$ by enabling it to validate errors, roll back to a previous state, and self-correct its mistakes~\cite{zhang2025enhancingwebagentsexplicit}. 
This approach differs from \textit{Reflection \& Lookahead} in that we explicitly focus on \textit{going back} to a previous state, whereas \textit{Reflection \& Lookahead} is more concerned with refining erroneous trajectories.

The core idea behind rollback is to equip $\mathcal{M}$ with the ability to dynamically evaluate the validity of its actions and their consequences at each step of the trajectory. 
When the outcome of an action deviates from $\mathcal{M}$'s plans or expectations, the model identifies the erroneous action $a_t$ and the corresponding state $o_t$ where the trajectory began to diverge. 
Subsequently, it rolls back to an earlier valid state $o_{t-1}$ and re-initiates the reasoning process from that point onward. 
Such mechanisms are particularly useful in tasks where irreversible errors can significantly impact the overall performance, such as web-based navigation or multi-step reasoning tasks.

To implement \textbf{R}ollback, we first randomly sample a successful trajectory, denoted as $\tau^R$, from the set of collected successful trajectories in $\mathcal{D}, \mathcal{D}^L, \text{and } \mathcal{D}^B$. 
Next, we randomly select \( n \) states, \( \{(o_j, h_j, a_j)\}^n_{j=1} \), from the sampled trajectory \( \tau^R \). For each observation \( o_j \), we generate an alternative thought \( h'_j \) and action \( a'_j \), ensuring that \( a'_j \neq a_j \). 
This is achieved using the prompt provided in Table~\ref{tab:alternative_actions_prompt}. 
We then roll out the corresponding outcome \( o'_{j+1} \) in a real web navigation environment, ensuring that \( o'_{j+1} \neq o_{j+1} \), such that there is a high chance that the new observation will lead to failure.
If the rollout actually lead to failure, we then regard the optimal strategy for \( o'_{j+1} \) as reverting to the previous observation \( o_j \) using the \textit{goback} action.
Once \( o'_{j+1} \) is determined as the state that requires \textit{goback}, we construct the corresponding thought \( h'_{j+1} \) using the prompt specified in Table~\ref{tab:backtracking_prompt}. 
Consequently, we create new rollback trajectories in the form: $\tau^{R'} = \{\cdots, (o_j, h_j, a'_j), (o'_{j+1}, h'_{j+1}, \mathrm{goback})\}$.
These rollback trajectories are then aggregated into the rollback training set  \( \mathcal{D}^{R} = \{(q, \tau^{R'})\}\).

\subsection{Cumulative Training}
\label{sec:cumu_training}
Instead of directly finetuning on all the trajectories sampled by the reasoning algorithms, 
we adopt a cumulative training strategy~\citep{bengio2009curriculum} to improve learning effectiveness and prevent overthinking.
The basic idea is that if a query can be successfully executed through simple self-exploration, there is no need to apply more complex thinking.
Starting from the baseline training dataset $\mathcal{D}_{rej}$ acquired by simple rejection sampling, we progressively append new trajectories that are successfully executed by the three reasoning algorithms but fail during self-exploration.
Based on the execution difficulty, we tested three variations of data mixing:

\noindent (1) Reflection \& \textbf{L}ookahead:

$\mathcal{D}_{\text{L}}^{c} = \mathcal{D}_{\text{rej}} \cup \left\{ (q, \tau) \in \mathcal{D}_L \;\middle|\; q \notin Q_{\text{rej}} \right\}$

\noindent (2) Reflection \& \textbf{L}ookahead + \textbf{B}ranching:

$\mathcal{D}_{\text{B}}^{c} = \mathcal{D}_{\text{L}}^c \cup \left\{ (q, \tau) \in \mathcal{D}_B \;\middle|\; q \notin Q_{\text{L}}^{c} \right\}$

\noindent (3) \textsc{WebCoT}: Reflection \& \textbf{L}ookahead + \textbf{B}ranching + \textbf{R}ollback:

$\mathcal{D}_{\text{R}}^{c} = \mathcal{D}_{\text{B}}^c \cup \left\{ (q, \tau) \in \mathcal{D}_R \;\middle|\; q \notin Q_{\text{B}}^c \right\}$

\begin{table*}[!ht]
    \centering
    \small
    \renewcommand{\arraystretch}{1.1}
    \setlength{\tabcolsep}{1.2mm}{
    \begin{tabular}{p{2.4cm}cccccccccccc}
    \toprule
         \multirow{3.5}{*}{Method}
         & \multicolumn{10}{c}{WebVoyager} & M2W & SimQA
         \\
         \cmidrule(lr){2-11} \cmidrule(lr){12-12} \cmidrule(lr){13-13} & \multirow{2}{*}{\tabincell{c}{ Apple}} & \multirow{2}{*}{\tabincell{c}{ ArXiv}} & \multirow{2}{*}{\tabincell{c}{ BBC}} & \multirow{2}{*}{ \tabincell{c}{ Cour-\\sera}} & \multirow{2}{*}{ \tabincell{c}{ \footnotesize{ESPN}}} & \multirow{2}{*}{ \tabincell{c}{ Git\\Hub}}
         & \multirow{2}{*}{ \tabincell{c}{ Google\\Map}} & \multirow{2}{*}{HF} & \multirow{2}{*}{ \tabincell{c}{ Wolfram\\Alpha}} & \multirow{2}{*}{ \tabincell{c}{ Avg.}} & \multirow{2}{*}{ \tabincell{c}{ Acc.}} & \multirow{2}{*}{ \tabincell{c}{ Acc.}} 
         \\
         \\
         \midrule
         \textsc{GPT-4o-mini} & 23.26 & 34.88 & 28.57 & 37.21 & 22.72 & 31.82 & 29.27 & 24.39 & 34.88 & 29.63 & 17.0 & 54.0\\
         \textsc{QwQ-32B}  & 27.91 & 11.63 & 38.10  & 38.10  & 29.55 & 17.07 & 48.78 & 20.93 & 52.17 & 31.69 & 15.1 & 27.0\\
         \textsc{GPT-4o}  & 30.23 & 20.93 & 28.57 & 51.16 & 30.95 & 38.64 & 24.39 & 29.27 & 56.52 & 34.54 & 18.8 & 61.0\\
        \midrule
        \midrule
        \textsc{Llama-3.3-70B} &  18.60 & \underline{23.26} & 23.81 & 23.81 & 15.91 & 26.83 & 31.71 & \textbf{30.23} & 28.26 & 24.68 & 5.7 & 25.0\\
        \textsc{+ Rej. Sampling}  &  \underline{34.88} & 6.98 & \underline{24.39} & 38.10 & 22.73  & \underline{34.15} & \textbf{48.78} & 19.05 & \underline{36.96} & 29.50 & 7.5 & 33.0\\
        \textsc{+ QwQ Distill}  & 32.56 & 20.93 & 23.81 & \underline{40.48} & \underline{29.55} & \underline{34.15} & 31.71 & 25.58 & \underline{36.96} & \underline{30.65} & \underline{18.9} & \underline{52.0}\\
         \textsc{+ \textbf{WebCoT}} & \textbf{39.53} & \textbf{27.91} & \textbf{30.95} & \textbf{59.52} & \textbf{38.64} &  \textbf{43.90} & \underline{46.34} & \underline{27.91} & \textbf{54.35} & \textbf{41.04} & \textbf{20.8} & \textbf{56.0}\\
         \bottomrule
    \end{tabular}
    }
    \caption{Performance comparison across WebVoyager, Mind2Web-Live (M2W), and SimpleQA (SimQA). The highest values are \textbf{bolded}, and the second highest is \underline{underlined}. \textsc{WebCoT} shows significant improvements, with \textsc{Llama-3.3-70B} + \textsc{WebCoT} outperforming \textsc{GPT-4o}, despite the latter's stronger foundational capacity. Furthermore, \textsc{Llama-3.3-70B} + \textsc{WebCoT} surpasses + \textsc{QwQ Distill} by 10.4\% on WebVoyager, highlighting our approach's effectiveness in web-specific reasoning tasks.}
    \label{tab:main_exp_combined}
\end{table*}

\section{Experiments}
\subsection{Setup}
We use the web agent module of the \textit{Cognitive Kernel}~\cite{cognitive_kernel} framework to conduct experiments.
In this setup, the state space $\mathcal{S}$ encompasses the entire Internet, facilitated by Playwright. 
The action space consists of primitive browser operations, including \texttt{type}, \texttt{click}, \texttt{scroll}, \texttt{goback}, \texttt{stop}, and \texttt{restart}. The observation at time step $t$, $o_t$, corresponds to the accessibility tree of visible components in the virtual browser—effectively simulating the perceptual experience of a human navigating the web. 
The transition function $\mathcal{T}$ executes the selected atomic browser actions based on the current webpage state, updates the webpage state, and propagates changes to the next observation $o_{t+1}$. 
Execution errors (e.g., navigation timeouts) are captured and relayed back to the reasoning module for appropriate handling, continuing until the task is completed or a predefined step limit is reached.

For task evaluation, we define a reward function $\mathcal{R}$ that mitigates issues with potential false negatives in human-annotated step-wise comparisons~\cite{webcanvas}. Specifically, we employ GPT-4o for end-to-end task completion assessment, following the methodology of \citet{webvoyager}. This evaluation strategy is designed to accommodate the inherent variability in task trajectories, where multiple distinct action sequences can achieve the same objective. GPT-4o is provided the complete task trajectory and the original query $q$, and it outputs a binary score (0 or 1) indicating whether the task has been completed. Detailed prompts are presented in Table~\ref{tab:eval_gpt_prompt}.

For all of our experiments, the agent leverages \texttt{Llama-3.3-70B} as the backbone foundation model $\mathcal{M}$. We only use the training queries of OpenWebVoyager~\cite{openwebvoyager} for trajectory collection and agent finetuning. During rejection sampling, \texttt{Llama-3.3-70B} itself is used to evaluate whether the task has been successfully completed or not. More details regarding the agent system, including definitions of the atomic operations, system prompts, are detailed in Appendix~\ref{sec:webcot_implementation}.

We evaluate the agent on three live web navigation benchmarks: {WebVoyager}~\cite{webvoyager}, {Mind2Web-Live}~\cite{webcanvas}, and {SimpleQA}~\cite{simpleqa}. These benchmarks require the web agent to interact with real-world web environments to complete a variety of tasks. To ensure experimental consistency, we filter out websites that are inaccessible due to geographical restrictions or IP blocks within our experimental setup (details in Appendix~\ref{sec:omitted_sites}). 

\subsection{Baselines}

In our experiments, we select four models as baselines for vanilla inference: GPT-4o-mini, GPT-4o, the advanced reasoning model QwQ-32B
, and Llama-3.3-70b-Instruct
. Additionally, we compare three data generation approaches for fine-tuning Llama-3.3-70b-Instruct: (i) Rejection Sampling (Rej. Sampling), (ii) successful trajectories sampled by \textsc{QwQ-32B} (QwQ Distill), and (iii) the \textsc{WebCoT} data proposed in our work for training.

\begin{table*}[!ht]
    \centering
    \small
    \renewcommand{\arraystretch}{1.1}
    \setlength{\tabcolsep}{1.2mm}{
    \begin{tabular}{p{2.4cm}cccccccccccc}
    \toprule
         \multirow{3.5}{*}{Method}
         & \multicolumn{10}{c}{WebVoyager} & M2W & SimQA \\
         \cmidrule(lr){2-11} \cmidrule(lr){12-12} \cmidrule(lr){13-13} & \multirow{2}{*}{\tabincell{c}{ Apple}} & \multirow{2}{*}{\tabincell{c}{ ArXiv}} & \multirow{2}{*}{\tabincell{c}{ BBC}} & \multirow{2}{*}{ \tabincell{c}{ Cour-\\sera}} & \multirow{2}{*}{ \tabincell{c}{ \footnotesize{ESPN}}} & \multirow{2}{*}{ \tabincell{c}{ Git\\Hub}}
         & \multirow{2}{*}{ \tabincell{c}{ Google\\Map}} & \multirow{2}{*}{HF} & \multirow{2}{*}{ \tabincell{c}{ Wolfram\\Alpha}} & \multirow{2}{*}{ \tabincell{c}{ Avg.}} & \multirow{2}{*}{ \tabincell{c}{ Acc.}} & \multirow{2}{*}{ \tabincell{c}{ Acc.}}\\
         \\
         \midrule
        \textsc{Llama-3.3} &  18.60 & \underline{23.26} & 23.81 & 23.81 & 15.91 & 26.83 & 31.71 & \underline{30.23} & 28.26 & 24.68 & 5.7 & 25.0\\
         + $\mathcal{D}_L^c$  & 30.23 & 9.30 & \textbf{38.10} & \underline{42.86} & 20.45 & \textbf{46.34} & 31.71 & 25.58 & 42.22 & 31.77 & \underline{17.0} & 49.0 \\
         + $\mathcal{D}^c_B$  & \underline{37.21} & \underline{18.60}
 & \underline{30.95} & 40.48 & \underline{36.36} & 34.15 & \underline{32.50} & \textbf{34.88} & \underline{43.48} & \underline{34.38} & 15.1 & \underline{52.0}\\
         \textsc{+ \textbf{WebCoT}} & \textbf{39.53} & \textbf{27.91} & \underline{30.95} & \textbf{59.52} & \textbf{38.64} &  \underline{43.90} & \textbf{46.34} & 27.91 & \textbf{54.35} & \textbf{41.04} & \textbf{20.8} & \textbf{56.0}\\
         \bottomrule
    \end{tabular}
    }
    \caption{Ablation study results on WebVoyager subtasks, M2W, and SimQA. The highest values are \textbf{bolded}, and the second highest values are \underline{underlined}. $\mathcal{D}_L^c$, $\mathcal{D}^c_B$, and \textsc{WebCoT} are detailed in Section~\ref{sec:cumu_training}. The results highlight the effectiveness of incorporating different reasoning components, with \textsc{WebCoT} showing the best performance across 3 benchmarks.}
    \label{tab:ablation_study}
\end{table*}
\subsection{Main Results}
Table~\ref{tab:main_exp_combined} presents the performance comparison between our method, \textsc{WebCoT}, and various baselines using the \textsc{Llama-3.3-70B} model across three benchmarks: {WebVoyager}, {Mind2Web-Live (M2W)}, and {SimpleQA (SimQA)}. The results clearly demonstrate that \textsc{WebCoT} achieves substantial improvements across all benchmarks. Specifically, the accuracy on WebVoyager increases by 16.5 points (a 66.8\% relative improvement). Similarly, it achieves gains of 15.1 points (264.9\%) and 31 points (124.0\%) on {M2W} and {SimQA}, respectively.

Notably, the capability of \textsc{WebCoT} even surpasses that of \textsc{GPT-4o}, which has significantly stronger foundational capacity compared to \textsc{Llama-3.3-70B}~\cite{achiam2023gpt,grattafiori2024llama}. These results underscore the potential of our approach in developing highly efficient and capable web agents. 

Compared to the baseline \textsc{QwQ Distill}, the systematically designed CoT in \textsc{WebCoT} demonstrates clear superiority. 
Unlike traditional reasoning models, which are primarily optimized for domains like mathematics and coding, \textsc{WebCoT} is specifically tailored to excel in reasoning chains for web-based tasks. 
This specialization makes it a more effective solution for navigating and reasoning within web environments. 
Notably, \textsc{WebCoT} outperforms \textsc{QwQ Distill} by 10.4 points on {WebVoyager}, highlighting its ability to better harness the potential of LLMs.

\subsection{Analysis}
\subsubsection{Effects of Different Reasoning Ability}
Table~\ref{tab:ablation_study} presents the performance improvements achieved by incorporating the training datasets created from each reasoning component: \textit{Reflection \& Lookahead}, \textit{Branching}, and \textit{Rollback}. The dataset $\mathcal{D}^R$ demonstrates a significant enhancement in the reasoning capabilities of LLMs. 

When comparing the use of $\mathcal{D}^c_L$
alone with the combined dataset $\mathcal{D}^c_B$
, we observe a discernible improvement across most benchmarks. This highlights the importance of instilling branching reasoning into LLMs. Furthermore, adding $\mathcal{D}_R$ to $\mathcal{D}^c_B$ yields even greater performance gains, highlighting the value of equipping LLMs with robust error validation capabilities to enable more efficient task completion within limited attempts.

\begin{table}[!ht]
    \centering
    \small
    \renewcommand{\arraystretch}{1.1}
    \setlength{\tabcolsep}{1.2mm}{
    \begin{tabular}{p{2cm}ccc}
    \toprule
         \multirow{1}{*}{Method}
         & WebVoyager & M2W & SimQA \\
         \midrule
         + $\mathcal{D}^c_L$  & \textbf{31.77} & \textbf{17.0} & \textbf{49.0} \\
         \textsc{Vanilla CoT} & {31.43} & {11.3} & {47.0}\\
         \midrule
          + $\mathcal{D}^c_B$ & \textbf{34.38} & {15.1} & \textbf{52.0}\\
         \textsc{Vanilla CoT} & {28.01} & \textbf{20.8} & {48.0}\\
         \bottomrule
    \end{tabular}
    }
    \caption{Ablation study on the effect of verbalizing new reasoning chain-of-thought versus using the original self-generated chain-of-thought (Vanilla CoT). Using the newly verbalized CoT will lead to more improvements in general.}
    \label{tab:cot_ablation}
\end{table}

\subsubsection{Effects of Rationale Verbalization}

In Reflection \& Lookahead and Branching, newly successful queries are executed and their corresponding trajectories are added to the training set. 
In \textsc{WebCoT}, we further verbalize lookahead planning and action selection as natural language rationales. 
To assess the impact of rationale verbalization, we conduct an ablation study: we compare finetuning the model with the new trajectories using either the original self-generated chain-of-thought or the \textsc{WebCoT} rationales, to determine which contributes more to performance improvements.
Specifically, we replaced all refined reasoning processes in $\mathcal{D}^c_L$
and $\mathcal{D}^c_B$ with their original, self-generated versions. Based on the \textsc{Llama-3.3-70B} checkpoint, we fine-tuned two corresponding variants of \textsc{Vanilla CoT}. The results are shown in Table~\ref{tab:cot_ablation}.

For $\mathcal{D}^c_L$
, the performance degradation in its corresponding \textsc{Vanilla CoT} variant is relatively small on the {WebVoyager} benchmark. This indicates that the trajectory refinement in $\mathcal{D}^L$ plays a critical role in boosting performance. However, the significant performance drops on {M2W} and {SimQA} further underscore the importance of reasoning process refinement in improving overall LLM abilities.

For the combined dataset $D^c_B$,
the impact of removing reasoning refinement is even more pronounced. The performance gap is substantial, with improvements of 6.4 points on {WebVoyager } benchmark and 4.0 points on {SimQA}. These results emphasize the necessity of refining the reasoning process to maximize the model's performance.

\subsubsection{Effects of Rollback on Task Efficiency}
We evaluate the effect of rollback reasoning by comparing the task completion efficiency of agents fine-tuned with $\mathcal{D}^c_B$ (\textit{Reflection \& Lookahead + Branching}) and $\mathcal{D}^c_R$ (\textit{Reflection \& Lookahead + Branching + Rollback}). To ensure a fair comparison, we identify 97 queries from the {WebVoyager} benchmark that both agents successfully completed (out of 428 total queries).

For each query, we measure the trajectory length (i.e., the number of steps to complete the task) for both agents. To focus on differences, we exclude queries where the trajectory lengths are identical, leaving 59 cases. For each of these, we calculate the trajectory length difference as:
\begin{equation}
\Delta L_i = |\tau^{B}_i| - |\tau^{R}_i|,
\end{equation}
where $|\tau^{B}_i|$ and $|\tau^{R}_i|$ are the trajectory lengths for agents fine-tuned with $\mathcal{D}^c_B$ and $\mathcal{D}^c_R$, respectively. A negative $\Delta L_i$ means the $\mathcal{D}^c_R$ agent required fewer steps.
\begin{figure}[t]
    \centering
    \includegraphics[width=0.46\textwidth]{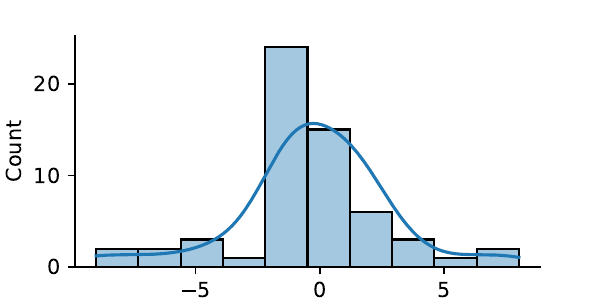} 
    \caption{Distribution of $\Delta L_i$. Including rollback mechanisms would lead to a reduced number of steps, indicating an improved action generation ability.}
    \label{fig:distri_delta_l}
    \vspace{-1em}
\end{figure}
Figure~\ref{fig:distri_delta_l} shows the distribution plot of $\Delta L_i$. On average, $\bar{\Delta L_i} = -0.23$, and the median is $\tilde{\Delta L_i} = -1.0$. These results indicate that the agent fine-tuned with $\mathcal{D}^c_R$ consistently completes tasks with fewer steps compared to the agent fine-tuned with $\mathcal{D}^c_B$. This highlights the efficiency improvement introduced by Rollback reasoning, demonstrating its ability to reduce unnecessary attempts and optimize task execution.

\begin{table}[t]
    \centering
    \small
    \renewcommand{\arraystretch}{1.1}
    \setlength{\tabcolsep}{1.2mm}{
    \begin{tabular}{p{2cm}ccc}
    \toprule
         \multirow{1}{*}{Method}
         & WebVoyager & M2W & SimQA \\
         \midrule
         + $\mathcal{D}^c_L$  & \textbf{31.77} & \textbf{17.0} & \textbf{49.0} \\
         + $\hat{\mathcal{D}}^c_{L}$ & 19.22 & 9.4 & 49.0\\
         \bottomrule
    \end{tabular}
    }
    \caption{Ablation on the effect of cumulative training. We fine-tune the LLM on $\hat{\mathcal{D}}^c_L$, which applies look-ahead planning CoT to all input queries rather than only the flawed ones, and find that this leads to degraded performance. }
    \label{tab:overthink_exp}
    \vspace{-0.6em}
\end{table}

\subsubsection{Effects of Cumulative Training}
To validate our design choice of \textit{Cumulative Training}, we conduct the following experiment: we construct a variant dataset, $\hat{\mathcal{D}}^c_L$, by refining all reasoning processes $h$ in trajectories—both with and without loops—from $\mathcal{D}_{rej}$. In contrast, $\mathcal{D}^c_L$ adheres to the cumulative training principle outlined in Section~\ref{sec:cumu_training}, refining $h$ only in trajectories with loops while leaving those without loops unchanged. The results, shown in Table~\ref{tab:overthink_exp}, demonstrate a significant performance drop across nearly all benchmarks when using $\hat{\mathcal{D}}^c_L$. Furthermore, hallucination behavior is frequently observed in the trajectories of \textsc{Llama-3.3} fine-tuned by $\hat{\mathcal{D}}^c_{L}$ (detailed in the Appendix~\ref{sec:hallu_case_study}). These findings further validate the effectiveness of our proposed design.
\begin{figure}[t]
    \centering
    \includegraphics[width=0.46\textwidth]{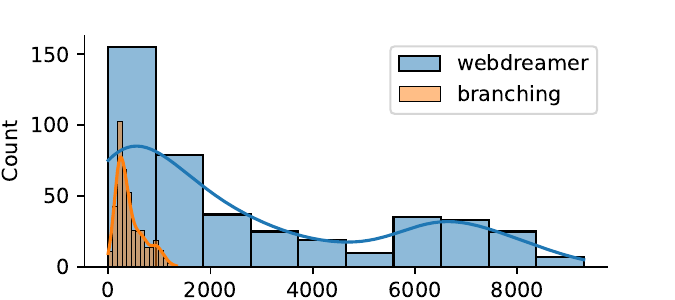} 
    \vspace{-1em}
    \caption{Distribution of generated token number per query for WebDreamer (branching) and \textsc{WebCoT}. A significantly smaller number of reasoning tokens is required for our method.}
    \label{fig:br_wd_diff_histogram}
    \vspace{-1em}
\end{figure}
\subsubsection{Token Efficiency of \textsc{WebCoT}}
Figure~\ref{fig:br_wd_diff_histogram} shows the distribution of tokens generated per query for \textsc{WebCoT} and WebDreamer on the WebVoyager test set. WebCoT uses significantly fewer tokens (mean: 422.2, median: 332.0) compared to WebDreamer (mean: 2665.6, median: 1580.0), highlighting WebCoT's efficiency and lower computational overhead. These results demonstrate that WebCoT achieves strong performance with reduced resource requirements.

\section{Conclusion}
We presented \textsc{WebCoT}, a framework that enhances the reasoning ability of LLMs for web-based agent tasks through \textit{reflection \& lookahead}, \textit{branching}, and \textit{rollback}. 
By curating and fine-tuning reasoning trajectories, \textsc{WebCoT} can improve the efficiency and task completion accuracy by a large margin. 
Our experiments across benchmarks such as WebVoyager, Mind2Web-Live, and SimpleQA show significant performance gains over baselines, 
including GPT-4o and QwQ-32B. 
These results highlight the value of targeted reasoning ability enhancements in bridging the gap between human and machine web agents. 

\section*{Limitations}
Our work focuses on enhancing web agent reasoning through explicit reflection, branching, and rollback mechanisms, but it does not include a comparison with Reinforcement Learning (RL)-based approaches. This omission is primarily due to the challenges associated with real-world web interactions, which are time-intensive and computationally expensive to simulate. Furthermore, real-world web environment rollouts are inherently non-deterministic, with variations in website behavior, latency, and accessibility affecting the outcome of RL experiments. These factors make direct comparisons with RL approaches infeasible within the scope of this study. Future work could explore hybrid methods that combine our structured reasoning framework with RL to further enhance web agent performance.

\section*{Ethics Statement}
This study was conducted in strict adherence to community ethical guidelines. The web agent datasets and benchmarks utilized in our research are documented as being safe, free from discrimination, personally identifiable information, and other potentially harmful content.
Additionally, we took great care in curating our instructions to the LLMs, ensuring that tasks were strictly confined to web navigation and excluded any activities that could raise ethical concerns. 

\section*{Acknowledgement}
Two authors (i.e., Minda Hu, Irwin King) of the work described in this paper were partially supported by the Research Grants Council of the Hong Kong Special Administrative Region, China (CUHK 2410072, RGC R1015-23). As the first author, I would like to express my heartfelt gratitude to my family, co-authors, and advisor, Prof. Irwin King, for their unwavering support and invaluable guidance throughout this work.


\bibliography{custom, agent_ref, agent_self_improve}

\begin{thebibliography}{53}
\providecommand{\natexlab}[1]{#1}

\bibitem[{Achiam et~al.(2023)Achiam, Adler, Agarwal, Ahmad, Akkaya, Aleman, Almeida, Altenschmidt, Altman, Anadkat et~al.}]{achiam2023gpt}
Josh Achiam, Steven Adler, Sandhini Agarwal, Lama Ahmad, Ilge Akkaya, Florencia~Leoni Aleman, Diogo Almeida, Janko Altenschmidt, Sam Altman, Shyamal Anadkat, and 1 others. 2023.
\newblock Gpt-4 technical report.
\newblock \emph{arXiv preprint arXiv:2303.08774}.

\bibitem[{Aksitov et~al.(2023)Aksitov, Miryoosefi, Li, Li, Babayan, Kopparapu, Fisher, Guo, Prakash, Srinivasan, Zaheer, Yu, and Kumar}]{rest_meets_react}
Renat Aksitov, Sobhan Miryoosefi, Zonglin Li, Daliang Li, Sheila Babayan, Kavya Kopparapu, Zachary Fisher, Ruiqi Guo, Sushant Prakash, Pranesh Srinivasan, Manzil Zaheer, Felix~X. Yu, and Sanjiv Kumar. 2023.
\newblock \href {https://doi.org/10.48550/ARXIV.2312.10003} {Rest meets react: Self-improvement for multi-step reasoning {LLM} agent}.
\newblock \emph{CoRR}, abs/2312.10003.

\bibitem[{Anthropic(2025)}]{claude37}
Anthropic. 2025.
\newblock Claude 3.7 sonnet: Hybrid reasoning model.
\newblock \url{https://www.anthropic.com/news/claude-3-7-sonnet}.
\newblock Accessed: 2025-04-18.

\bibitem[{Bengio et~al.(2009)Bengio, Louradour, Collobert, and Weston}]{bengio2009curriculum}
Yoshua Bengio, J{\'e}r{\^o}me Louradour, Ronan Collobert, and Jason Weston. 2009.
\newblock Curriculum learning.
\newblock In \emph{Proceedings of the 26th annual international conference on machine learning}, pages 41--48.

\bibitem[{Chen et~al.(2024)Chen, Saha, Stengel-Eskin, and Bansal}]{chen2024magdi}
Justin Chih-Yao Chen, Swarnadeep Saha, Elias Stengel-Eskin, and Mohit Bansal. 2024.
\newblock Magdi: Structured distillation of multi-agent interaction graphs improves reasoning in smaller language models.
\newblock \emph{arXiv preprint arXiv:2402.01620}.

\bibitem[{Cuadron et~al.(2025)Cuadron, Li, Ma, Wang, Wang, Zhuang, Liu, Schroeder, Xia, Mao, Thumiger, Desai, Stoica, Klimovic, Neubig, and Gonzalez}]{DBLP:journals/corr/abs-2502-08235}
Alejandro Cuadron, Dacheng Li, Wenjie Ma, Xingyao Wang, Yichuan Wang, Siyuan Zhuang, Shu Liu, Luis~Gaspar Schroeder, Tian Xia, Huanzhi Mao, Nicholas Thumiger, Aditya Desai, Ion Stoica, Ana Klimovic, Graham Neubig, and Joseph~E. Gonzalez. 2025.
\newblock \href {https://doi.org/10.48550/ARXIV.2502.08235} {The danger of overthinking: Examining the reasoning-action dilemma in agentic tasks}.
\newblock \emph{CoRR}, abs/2502.08235.

\bibitem[{Dang and Ngo(2025)}]{dang2025reinforcement}
Quy-Anh Dang and Chris Ngo. 2025.
\newblock Reinforcement learning for reasoning in small llms: What works and what doesn't.
\newblock \emph{arXiv preprint arXiv:2503.16219}.

\bibitem[{DeepSeek-AI et~al.(2025)DeepSeek-AI, Guo, Yang, Zhang, Song, Zhang, Xu, Zhu, Ma, Wang, Bi, Zhang, Yu, Wu, Wu, Gou, Shao, Li, Gao, Liu, Xue, Wang, Wu, Feng, Lu, Zhao, Deng, Zhang, Ruan, Dai, Chen, Ji, Li, Lin, Dai, Luo, Hao, Chen, Li, Zhang, Bao, Xu, Wang, Ding, Xin, Gao, Qu, Li, Guo, Li, Wang, Chen, Yuan, Qiu, Li, Cai, Ni, Liang, Chen, Dong, Hu, Gao, Guan, Huang, Yu, Wang, Zhang, Zhao, Wang, Zhang, Xu, Xia, Zhang, Zhang, Tang, Li, Wang, Li, Tian, Huang, Zhang, Wang, Chen, Du, Ge, Zhang, Pan, Wang, Chen, Jin, Chen, Lu, Zhou, Chen, Ye, Wang, Yu, Zhou, Pan, Li, Zhou, Wu, Ye, Yun, Pei, Sun, Wang, Zeng, Zhao, Liu, Liang, Gao, Yu, Zhang, Xiao, An, Liu, Wang, Chen, Nie, Cheng, Liu, Xie, Liu, Yang, Li, Su, Lin, Li, Jin, Shen, Chen, Sun, Wang, Song, Zhou, Wang, Shan, Li, Wang, Wei, Zhang, Xu, Li, Zhao, Sun, Wang, Yu, Zhang, Shi, Xiong, He, Piao, Wang, Tan, Ma, Liu, Guo, Ou, Wang, Gong, Zou, He, Xiong, Luo, You, Liu, Zhou, Zhu, Xu, Huang, Li, Zheng, Zhu, Ma, Tang, Zha, Yan, Ren, Ren, Sha, Fu, Xu, Xie, Zhang,
  Hao, Ma, Yan, Wu, Gu, Zhu, Liu, Li, Xie, Song, Pan, Huang, Xu, Zhang, and Zhang}]{deepseekr1}
DeepSeek-AI, Daya Guo, Dejian Yang, Haowei Zhang, Junxiao Song, Ruoyu Zhang, Runxin Xu, Qihao Zhu, Shirong Ma, Peiyi Wang, Xiao Bi, Xiaokang Zhang, Xingkai Yu, Yu~Wu, Z.~F. Wu, Zhibin Gou, Zhihong Shao, Zhuoshu Li, Ziyi Gao, and 181 others. 2025.
\newblock \href {https://arxiv.org/abs/2501.12948} {Deepseek-r1: Incentivizing reasoning capability in llms via reinforcement learning}.
\newblock \emph{Preprint}, arXiv:2501.12948.

\bibitem[{Dubey et~al.(2024)Dubey, Jauhri, Pandey, Kadian, Al{-}Dahle, Letman, Mathur, Schelten, Yang, Fan, Goyal, Hartshorn, Yang, Mitra, Sravankumar, Korenev, Hinsvark, Rao, Zhang, Rodriguez, Gregerson, Spataru, Rozi{\`{e}}re, Biron, Tang, Chern, Caucheteux, Nayak, Bi, Marra, McConnell, Keller, Touret, Wu, Wong, Ferrer, Nikolaidis, Allonsius, Song, Pintz, Livshits, Esiobu, Choudhary, Mahajan, Garcia{-}Olano, Perino, Hupkes, Lakomkin, AlBadawy, Lobanova, Dinan, Smith, Radenovic, Zhang, Synnaeve, Lee, Anderson, Nail, Mialon, Pang, Cucurell, Nguyen, Korevaar, Xu, Touvron, Zarov, Ibarra, Kloumann, Misra, Evtimov, Copet, Lee, Geffert, Vranes, Park, Mahadeokar, Shah, van~der Linde, Billock, Hong, Lee, Fu, Chi, Huang, Liu, Wang, Yu, Bitton, Spisak, Park, Rocca, Johnstun, Saxe, Jia, Alwala, Upasani, Plawiak, Li, Heafield, Stone, and et~al.}]{llama3}
Abhimanyu Dubey, Abhinav Jauhri, Abhinav Pandey, Abhishek Kadian, Ahmad Al{-}Dahle, Aiesha Letman, Akhil Mathur, Alan Schelten, Amy Yang, Angela Fan, Anirudh Goyal, Anthony Hartshorn, Aobo Yang, Archi Mitra, Archie Sravankumar, Artem Korenev, Arthur Hinsvark, Arun Rao, Aston Zhang, and 82 others. 2024.
\newblock \href {https://doi.org/10.48550/ARXIV.2407.21783} {The llama 3 herd of models}.
\newblock \emph{CoRR}, abs/2407.21783.

\bibitem[{Fang et~al.(2025{\natexlab{a}})Fang, Zhang, Zhang, Ma, Yu, Mi, and Yu}]{fang2025webevolver}
Tianqing Fang, Hongming Zhang, Zhisong Zhang, Kaixin Ma, Wenhao Yu, Haitao Mi, and Dong Yu. 2025{\natexlab{a}}.
\newblock Webevolver: Enhancing web agent self-improvement with coevolving world model.
\newblock \emph{arXiv preprint arXiv:2504.21024}.

\bibitem[{Fang et~al.(2025{\natexlab{b}})Fang, Zhang, Wang, Wang, Qin, Wan, Ma, Zhang, Chen, Li et~al.}]{fang2025cognitive}
Tianqing Fang, Zhisong Zhang, Xiaoyang Wang, Rui Wang, Can Qin, Yuxuan Wan, Jun-Yu Ma, Ce~Zhang, Jiaqi Chen, Xiyun Li, and 1 others. 2025{\natexlab{b}}.
\newblock Cognitive kernel-pro: A framework for deep research agents and agent foundation models training.
\newblock \emph{arXiv preprint arXiv:2508.00414}.

\bibitem[{Grattafiori et~al.(2024)Grattafiori, Dubey, Jauhri, Pandey, Kadian, Al-Dahle, Letman, Mathur, Schelten, Vaughan et~al.}]{grattafiori2024llama}
Aaron Grattafiori, Abhimanyu Dubey, Abhinav Jauhri, Abhinav Pandey, Abhishek Kadian, Ahmad Al-Dahle, Aiesha Letman, Akhil Mathur, Alan Schelten, Alex Vaughan, and 1 others. 2024.
\newblock The llama 3 herd of models.
\newblock \emph{arXiv preprint arXiv:2407.21783}.

\bibitem[{Gu et~al.(2024)Gu, Zheng, Gou, Zhang, Chang, Srivastava, Xie, Qi, Sun, and Su}]{webdreamer}
Yu~Gu, Boyuan Zheng, Boyu Gou, Kai Zhang, Cheng Chang, Sanjari Srivastava, Yanan Xie, Peng Qi, Huan Sun, and Yu~Su. 2024.
\newblock \href {https://doi.org/10.48550/ARXIV.2411.06559} {Is your {LLM} secretly a world model of the internet? model-based planning for web agents}.
\newblock \emph{CoRR}, abs/2411.06559.

\bibitem[{He et~al.(2024{\natexlab{a}})He, Yao, Ma, Yu, Dai, Zhang, Lan, and Yu}]{webvoyager}
Hongliang He, Wenlin Yao, Kaixin Ma, Wenhao Yu, Yong Dai, Hongming Zhang, Zhenzhong Lan, and Dong Yu. 2024{\natexlab{a}}.
\newblock \href {https://doi.org/10.18653/V1/2024.ACL-LONG.371} {Webvoyager: Building an end-to-end web agent with large multimodal models}.
\newblock In \emph{Proceedings of the 62nd Annual Meeting of the Association for Computational Linguistics (Volume 1: Long Papers), {ACL} 2024, Bangkok, Thailand, August 11-16, 2024}, pages 6864--6890. Association for Computational Linguistics.

\bibitem[{He et~al.(2024{\natexlab{b}})He, Yao, Ma, Yu, Zhang, Fang, Lan, and Yu}]{openwebvoyager}
Hongliang He, Wenlin Yao, Kaixin Ma, Wenhao Yu, Hongming Zhang, Tianqing Fang, Zhenzhong Lan, and Dong Yu. 2024{\natexlab{b}}.
\newblock \href {https://doi.org/10.48550/ARXIV.2410.19609} {Openwebvoyager: Building multimodal web agents via iterative real-world exploration, feedback and optimization}.
\newblock \emph{CoRR}, abs/2410.19609.

\bibitem[{Hu et~al.(2024)Hu, Zong, Wang, Zhou, Li, Gao, Wong, Li, and King}]{hu-etal-2024-serts}
Minda Hu, Licheng Zong, Hongru Wang, Jingyan Zhou, Jingjing Li, Yichen Gao, Kam-Fai Wong, Yu~Li, and Irwin King. 2024.
\newblock \href {https://doi.org/10.18653/v1/2024.findings-emnlp.71} {{S}e{RTS}: Self-rewarding tree search for biomedical retrieval-augmented generation}.
\newblock In \emph{Findings of the Association for Computational Linguistics: EMNLP 2024}, pages 1321--1335, Miami, Florida, USA. Association for Computational Linguistics.

\bibitem[{Hu et~al.(2025)Hu, Wang, Dong, Xu, Saha, Xiong, Hooi, and Li}]{hu2025ahasystematicmetaabilitiesalignment}
Zhiyuan Hu, Yibo Wang, Hanze Dong, Yuhui Xu, Amrita Saha, Caiming Xiong, Bryan Hooi, and Junnan Li. 2025.
\newblock \href {https://arxiv.org/abs/2505.10554} {Beyond 'aha!': Toward systematic meta-abilities alignment in large reasoning models}.
\newblock \emph{Preprint}, arXiv:2505.10554.

\bibitem[{Jia et~al.(2024)Jia, Yu, Ma, Fang, Zhang, Ouyang, Zhang, Jiang, and Yu}]{leopard}
Mengzhao Jia, Wenhao Yu, Kaixin Ma, Tianqing Fang, Zhihan Zhang, Siru Ouyang, Hongming Zhang, Meng Jiang, and Dong Yu. 2024.
\newblock \href {https://doi.org/10.48550/ARXIV.2410.01744} {Leopard: {A} vision language model for text-rich multi-image tasks}.
\newblock \emph{CoRR}, abs/2410.01744.

\bibitem[{Kumar et~al.(2025)Kumar, Roh, Naseh, Karpinska, Iyyer, Houmansadr, and Bagdasarian}]{kumar2025overthinking}
Abhinav Kumar, Jaechul Roh, Ali Naseh, Marzena Karpinska, Mohit Iyyer, Amir Houmansadr, and Eugene Bagdasarian. 2025.
\newblock Overthinking: Slowdown attacks on reasoning llms.
\newblock \emph{arXiv preprint arXiv:2502.02542}.

\bibitem[{Li et~al.(2025)Li, Jin, Dong, Qian, Zhu, Wu, Wen, and Dou}]{webthinker}
Xiaoxi Li, Jiajie Jin, Guanting Dong, Hongjin Qian, Yutao Zhu, Yongkang Wu, Ji-Rong Wen, and Zhicheng Dou. 2025.
\newblock \href {https://arxiv.org/abs/2504.21776} {Webthinker: Empowering large reasoning models with deep research capability}.
\newblock \emph{Preprint}, arXiv:2504.21776.

\bibitem[{Liang et~al.(2025)Liang, Xiang, Yu, Zhang, and Hong}]{openmanus2025}
Xinbin Liang, Jinyu Xiang, Zhaoyang Yu, Jiayi Zhang, and Sirui Hong. 2025.
\newblock Openmanus: An open-source framework for building general ai agents.
\newblock \url{https://github.com/mannaandpoem/OpenManus}.

\bibitem[{Liu et~al.(2025)Liu, Li, Xie, Hu, Han, Zhang, Yang, and Wu}]{liu2025infigui}
Yuhang Liu, Pengxiang Li, Congkai Xie, Xavier Hu, Xiaotian Han, Shengyu Zhang, Hongxia Yang, and Fei Wu. 2025.
\newblock Infigui-r1: Advancing multimodal gui agents from reactive actors to deliberative reasoners.
\newblock \emph{arXiv preprint arXiv:2504.14239}.

\bibitem[{Mialon et~al.(2024)Mialon, Fourrier, Wolf, LeCun, and Scialom}]{GAIA}
Gr{\'{e}}goire Mialon, Cl{\'{e}}mentine Fourrier, Thomas Wolf, Yann LeCun, and Thomas Scialom. 2024.
\newblock \href {https://openreview.net/forum?id=fibxvahvs3} {{GAIA:} a benchmark for general {AI} assistants}.
\newblock In \emph{The Twelfth International Conference on Learning Representations, {ICLR} 2024, Vienna, Austria, May 7-11, 2024}. OpenReview.net.

\bibitem[{{Monica.Im}(2025)}]{manus}
{Monica.Im}. 2025.
\newblock \href {https://manus.im/} {Manus ai}.
\newblock Technical report, Monica.Im.

\bibitem[{OpenAI et~al.(2024)OpenAI, :, Jaech, Kalai, Lerer, Richardson, El-Kishky, Low, Helyar, Madry, Beutel, Carney, Iftimie, Karpenko, Passos, Neitz, Prokofiev, Wei, Tam, Bennett, Kumar, Saraiva, Vallone, Duberstein, Kondrich, Mishchenko, Applebaum, Jiang, Nair, Zoph, Ghorbani, Rossen, Sokolowsky, Barak, McGrew, Minaiev, Hao, Baker, Houghton, McKinzie, Eastman, Lugaresi, Bassin, Hudson, Li, de~Bourcy, Voss, Shen, Zhang, Koch, Orsinger, Hesse, Fischer, Chan, Roberts, Kappler, Levy, Selsam, Dohan, Farhi, Mely, Robinson, Tsipras, Li, Oprica, Freeman, Zhang, Wong, Proehl, Cheung, Mitchell, Wallace, Ritter, Mays, Wang, Such, Raso, Leoni, Tsimpourlas, Song, von Lohmann, Sulit, Salmon, Parascandolo, Chabot, Zhao, Brockman, Leclerc, Salman, Bao, Sheng, Andrin, Bagherinezhad, Ren, Lightman, Chung, Kivlichan, O'Connell, Osband, Gilaberte, Akkaya, Kostrikov, Sutskever, Kofman, Pachocki, Lennon, Wei, Harb, Twore, Feng, Yu, Weng, Tang, Yu, Candela, Palermo, Parish, Heidecke, Hallman, Rizzo, Gordon, Uesato, Ward,
  Huizinga, Wang, Chen, Xiao, Singhal, Nguyen, Cobbe, Shi, Wood, Rimbach, Gu-Lemberg, Liu, Lu, Stone, Yu, Ahmad, Yang, Liu, Maksin, Ho, Fedus, Weng, Li, McCallum, Held, Kuhn, Kondraciuk, Kaiser, Metz, Boyd, Trebacz, Joglekar, Chen, Tintor, Meyer, Jones, Kaufer, Schwarzer, Shah, Yatbaz, Guan, Xu, Yan, Glaese, Chen, Lampe, Malek, Wang, Fradin, McClay, Pavlov, Wang, Wang, Murati, Bavarian, Rohaninejad, McAleese, Chowdhury, Chowdhury, Ryder, Tezak, Brown, Nachum, Boiko, Murk, Watkins, Chao, Ashbourne, Izmailov, Zhokhov, Dias, Arora, Lin, Lopes, Gaon, Miyara, Leike, Hwang, Garg, Brown, James, Shu, Cheu, Greene, Jain, Altman, Toizer, Toyer, Miserendino, Agarwal, Hernandez, Baker, McKinney, Yan, Zhao, Hu, Santurkar, Chaudhuri, Zhang, Fu, Papay, Lin, Balaji, Sanjeev, Sidor, Broda, Clark, Wang, Gordon, Sanders, Patwardhan, Sottiaux, Degry, Dimson, Zheng, Garipov, Stasi, Bansal, Creech, Peterson, Eloundou, Qi, Kosaraju, Monaco, Pong, Fomenko, Zheng, Zhou, McCabe, Zaremba, Dubois, Lu, Chen, Cha, Bai, He, Zhang, Wang,
  Shao, and Li}]{openai2024openaio1card}
OpenAI, :, Aaron Jaech, Adam Kalai, Adam Lerer, Adam Richardson, Ahmed El-Kishky, Aiden Low, Alec Helyar, Aleksander Madry, Alex Beutel, Alex Carney, Alex Iftimie, Alex Karpenko, Alex~Tachard Passos, Alexander Neitz, Alexander Prokofiev, Alexander Wei, Allison Tam, and 244 others. 2024.
\newblock \href {https://arxiv.org/abs/2412.16720} {Openai o1 system card}.
\newblock \emph{Preprint}, arXiv:2412.16720.

\bibitem[{OpenAI(2023)}]{gpt4}
OpenAI. 2023.
\newblock \href {https://arxiv.org/abs/2303.08774} {Gpt-4 technical report}.
\newblock Technical Report.
\newblock A large multimodal model capable of processing image and text inputs and producing text outputs. Achieves human-level performance on various professional benchmarks including passing a simulated bar exam in the top 10

\bibitem[{Pahuja et~al.(2025)Pahuja, Lu, Rosset, Gou, Mitra, Whitehead, Su, and Awadallah}]{Explorer}
Vardaan Pahuja, Yadong Lu, Corby Rosset, Boyu Gou, Arindam Mitra, Spencer Whitehead, Yu~Su, and Ahmed Awadallah. 2025.
\newblock \href {https://doi.org/10.48550/ARXIV.2502.11357} {Explorer: Scaling exploration-driven web trajectory synthesis for multimodal web agents}.
\newblock \emph{CoRR}, abs/2502.11357.

\bibitem[{Pan et~al.(2024)Pan, Kong, Zhou, Cui, Leng, Jiang, Liu, Shang, Zhou, Wu, and Wu}]{webcanvas}
Yichen Pan, Dehan Kong, Sida Zhou, Cheng Cui, Yifei Leng, Bing Jiang, Hangyu Liu, Yanyi Shang, Shuyan Zhou, Tongshuang Wu, and Zhengyang Wu. 2024.
\newblock \href {https://doi.org/10.48550/ARXIV.2406.12373} {Webcanvas: Benchmarking web agents in online environments}.
\newblock \emph{CoRR}, abs/2406.12373.

\bibitem[{Patel et~al.(2024)Patel, Hofmarcher, Leoveanu{-}Condrei, Dinu, Callison{-}Burch, and Hochreiter}]{LLM_agent_can_self_improve}
Ajay Patel, Markus Hofmarcher, Claudiu Leoveanu{-}Condrei, Marius{-}Constantin Dinu, Chris Callison{-}Burch, and Sepp Hochreiter. 2024.
\newblock \href {https://doi.org/10.48550/ARXIV.2405.20309} {Large language models can self-improve at web agent tasks}.
\newblock \emph{CoRR}, abs/2405.20309.

\bibitem[{Qi et~al.(2025)Qi, Liu, Iong, Lai, Sun, Zhao, Yang, Yang, Sun, Yao, Zhang, Xu, Tang, and Dong}]{webrl}
Zehan Qi, Xiao Liu, Iat~Long Iong, Hanyu Lai, Xueqiao Sun, Wenyi Zhao, Yu~Yang, Xinyue Yang, Jiadai Sun, Shuntian Yao, Tianjie Zhang, Wei Xu, Jie Tang, and Yuxiao Dong. 2025.
\newblock \href {https://arxiv.org/abs/2411.02337} {Webrl: Training llm web agents via self-evolving online curriculum reinforcement learning}.
\newblock \emph{Preprint}, arXiv:2411.02337.

\bibitem[{Qwen(2025)}]{qwen2025qwq32b}
Qwen. 2025.
\newblock Qwq-32b: A compact reasoning model with reinforcement learning scaling.
\newblock \url{https://huggingface.co/Qwen/QwQ-32B}.
\newblock Apache 2.0 License. Model available at \url{https://huggingface.co/Qwen/QwQ-32B}.

\bibitem[{Shinn et~al.(2023)Shinn, Cassano, Gopinath, Narasimhan, and Yao}]{Reflexion}
Noah Shinn, Federico Cassano, Ashwin Gopinath, Karthik Narasimhan, and Shunyu Yao. 2023.
\newblock \href {http://papers.nips.cc/paper\_files/paper/2023/hash/1b44b878bb782e6954cd888628510e90-Abstract-Conference.html} {Reflexion: language agents with verbal reinforcement learning}.
\newblock In \emph{Advances in Neural Information Processing Systems 36: Annual Conference on Neural Information Processing Systems 2023, NeurIPS 2023, New Orleans, LA, USA, December 10 - 16, 2023}.

\bibitem[{Singh et~al.(2025)Singh, Magazine, Pandya, and Nambi}]{singh2025agentic}
Joykirat Singh, Raghav Magazine, Yash Pandya, and Akshay Nambi. 2025.
\newblock Agentic reasoning and tool integration for llms via reinforcement learning.
\newblock \emph{arXiv preprint arXiv:2505.01441}.

\bibitem[{Song et~al.(2025)Song, Thai, Pham, Chang, Nadaf, and Iyyer}]{song2025bearcubs}
Yixiao Song, Katherine Thai, Chau~Minh Pham, Yapei Chang, Mazin Nadaf, and Mohit Iyyer. 2025.
\newblock Bearcubs: A benchmark for computer-using web agents.
\newblock \emph{arXiv preprint arXiv:2503.07919}.

\bibitem[{Su et~al.(2025)Su, Healey, Nakov, and Cardie}]{su2025between}
Jinyan Su, Jennifer Healey, Preslav Nakov, and Claire Cardie. 2025.
\newblock Between underthinking and overthinking: An empirical study of reasoning length and correctness in llms.
\newblock \emph{arXiv preprint arXiv:2505.00127}.

\bibitem[{Sun et~al.(2025)Sun, Cheng, Ding, Jin, Wang, Xu, Wu, Jia, Chen, Liu, Kao, Li, He, Qiao, and Wu}]{sun2025osgenesisautomatingguiagent}
Qiushi Sun, Kanzhi Cheng, Zichen Ding, Chuanyang Jin, Yian Wang, Fangzhi Xu, Zhenyu Wu, Chengyou Jia, Liheng Chen, Zhoumianze Liu, Ben Kao, Guohao Li, Junxian He, Yu~Qiao, and Zhiyong Wu. 2025.
\newblock \href {https://arxiv.org/abs/2412.19723} {Os-genesis: Automating gui agent trajectory construction via reverse task synthesis}.
\newblock \emph{Preprint}, arXiv:2412.19723.

\bibitem[{Trabucco et~al.(2025)Trabucco, Sigurdsson, Piramuthu, and Salakhutdinov}]{DBLP:journals/corr/abs-2502-06776}
Brandon Trabucco, Gunnar~A. Sigurdsson, Robinson Piramuthu, and Ruslan Salakhutdinov. 2025.
\newblock \href {https://doi.org/10.48550/ARXIV.2502.06776} {Towards internet-scale training for agents}.
\newblock \emph{CoRR}, abs/2502.06776.

\bibitem[{Wei et~al.(2024)Wei, Karina, Chung, Jiao, Papay, Glaese, Schulman, and Fedus}]{simpleqa}
Jason Wei, Nguyen Karina, Hyung~Won Chung, Yunxin~Joy Jiao, Spencer Papay, Amelia Glaese, John Schulman, and William Fedus. 2024.
\newblock Measuring short-form factuality in large language models.
\newblock \emph{arXiv preprint arXiv:2411.04368}.

\bibitem[{Wei et~al.(2022)Wei, Wang, Schuurmans, Bosma, Xia, Chi, Le, Zhou et~al.}]{wei2022chain}
Jason Wei, Xuezhi Wang, Dale Schuurmans, Maarten Bosma, Fei Xia, Ed~Chi, Quoc~V Le, Denny Zhou, and 1 others. 2022.
\newblock Chain-of-thought prompting elicits reasoning in large language models.
\newblock \emph{Advances in neural information processing systems}, 35:24824--24837.

\bibitem[{Wei et~al.(2025)Wei, Yao, Liu, Zhang, Lu, Qiu, Yu, Xu, Zhang, Yin, Yun, and Li}]{wei2025webagentr1trainingwebagents}
Zhepei Wei, Wenlin Yao, Yao Liu, Weizhi Zhang, Qin Lu, Liang Qiu, Changlong Yu, Puyang Xu, Chao Zhang, Bing Yin, Hyokun Yun, and Lihong Li. 2025.
\newblock \href {https://arxiv.org/abs/2505.16421} {Webagent-r1: Training web agents via end-to-end multi-turn reinforcement learning}.
\newblock \emph{Preprint}, arXiv:2505.16421.

\bibitem[{Wu et~al.(2025{\natexlab{a}})Wu, Yin, Jiang, Wang, Xi, Fang, Zhang, He, Zhou, Xie, and Huang}]{WebWalker}
Jialong Wu, Wenbiao Yin, Yong Jiang, Zhenglin Wang, Zekun Xi, Runnan Fang, Linhai Zhang, Yulan He, Deyu Zhou, Pengjun Xie, and Fei Huang. 2025{\natexlab{a}}.
\newblock \href {https://doi.org/10.48550/ARXIV.2501.07572} {Webwalker: Benchmarking llms in web traversal}.
\newblock \emph{CoRR}, abs/2501.07572.

\bibitem[{Wu et~al.(2025{\natexlab{b}})Wu, Zhu, and Liu}]{wu2025agentic}
Junde Wu, Jiayuan Zhu, and Yuyuan Liu. 2025{\natexlab{b}}.
\newblock Agentic reasoning: Reasoning llms with tools for the deep research.
\newblock \emph{arXiv preprint arXiv:2502.04644}.

\bibitem[{Xu et~al.(2025)Xu, Hao, Zong, Wang, Zhang, Wang, Lan, Gong, Ouyang, Meng et~al.}]{xu2025towards}
Fengli Xu, Qianyue Hao, Zefang Zong, Jingwei Wang, Yunke Zhang, Jingyi Wang, Xiaochong Lan, Jiahui Gong, Tianjian Ouyang, Fanjin Meng, and 1 others. 2025.
\newblock Towards large reasoning models: A survey of reinforced reasoning with large language models.
\newblock \emph{arXiv preprint arXiv:2501.09686}.

\bibitem[{Xu et~al.(2024)Xu, Lu, Shen, Wang, Wang, Mao, Xiong, and Yu}]{xu2024agenttrek}
Yiheng Xu, Dunjie Lu, Zhennan Shen, Junli Wang, Zekun Wang, Yuchen Mao, Caiming Xiong, and Tao Yu. 2024.
\newblock Agenttrek: Agent trajectory synthesis via guiding replay with web tutorials.
\newblock \emph{arXiv preprint arXiv:2412.09605}.

\bibitem[{Yao et~al.(2022)Yao, Chen, Yang, and Narasimhan}]{webshop}
Shunyu Yao, Howard Chen, John Yang, and Karthik Narasimhan. 2022.
\newblock \href {http://papers.nips.cc/paper\_files/paper/2022/hash/82ad13ec01f9fe44c01cb91814fd7b8c-Abstract-Conference.html} {Webshop: Towards scalable real-world web interaction with grounded language agents}.
\newblock In \emph{Advances in Neural Information Processing Systems 35: Annual Conference on Neural Information Processing Systems 2022, NeurIPS 2022, New Orleans, LA, USA, November 28 - December 9, 2022}.

\bibitem[{Yao et~al.(2023)Yao, Zhao, Yu, Du, Shafran, Narasimhan, and Cao}]{yaoreact}
Shunyu Yao, Jeffrey Zhao, Dian Yu, Nan Du, Izhak Shafran, Karthik~R. Narasimhan, and Yuan Cao. 2023.
\newblock \href {https://openreview.net/forum?id=WE\_vluYUL-X} {React: Synergizing reasoning and acting in language models}.
\newblock In \emph{The Eleventh International Conference on Learning Representations, {ICLR} 2023, Kigali, Rwanda, May 1-5, 2023}. OpenReview.net.

\bibitem[{Zhang et~al.(2025{\natexlab{a}})Zhang, Hong, and Yu}]{zhang2025streaminglookingaheadtokenlevel}
Hongming Zhang, Ruixin Hong, and Dong Yu. 2025{\natexlab{a}}.
\newblock \href {https://arxiv.org/abs/2503.00029} {Streaming looking ahead with token-level self-reward}.
\newblock \emph{Preprint}, arXiv:2503.00029.

\bibitem[{Zhang et~al.(2024{\natexlab{a}})Zhang, Pan, Wang, Ma, Yu, and Yu}]{cognitive_kernel}
Hongming Zhang, Xiaoman Pan, Hongwei Wang, Kaixin Ma, Wenhao Yu, and Dong Yu. 2024{\natexlab{a}}.
\newblock \href {https://doi.org/10.48550/ARXIV.2409.10277} {Cognitive kernel: An open-source agent system towards generalist autopilots}.
\newblock \emph{CoRR}, abs/2409.10277.

\bibitem[{Zhang et~al.(2025{\natexlab{b}})Zhang, Yang, Shu, Wen, and Sang}]{zhang2025agent}
Yuxiang Zhang, Yuqi Yang, Jiangming Shu, Xinyan Wen, and Jitao Sang. 2025{\natexlab{b}}.
\newblock Agent models: Internalizing chain-of-action generation into reasoning models.
\newblock \emph{arXiv preprint arXiv:2503.06580}.

\bibitem[{Zhang et~al.(2025{\natexlab{c}})Zhang, Fang, Ma, Yu, Zhang, Mi, and Yu}]{zhang2025enhancingwebagentsexplicit}
Zhisong Zhang, Tianqing Fang, Kaixin Ma, Wenhao Yu, Hongming Zhang, Haitao Mi, and Dong Yu. 2025{\natexlab{c}}.
\newblock \href {https://arxiv.org/abs/2504.11788} {Enhancing web agents with explicit rollback mechanisms}.
\newblock \emph{Preprint}, arXiv:2504.11788.

\bibitem[{Zhang et~al.(2024{\natexlab{b}})Zhang, Tian, Chen, and Liu}]{MMInA}
Ziniu Zhang, Shulin Tian, Liangyu Chen, and Ziwei Liu. 2024{\natexlab{b}}.
\newblock \href {https://doi.org/10.48550/ARXIV.2404.09992} {Mmina: Benchmarking multihop multimodal internet agents}.
\newblock \emph{CoRR}, abs/2404.09992.

\bibitem[{Zhao et~al.(2024)Zhao, Ma, Chai, Wang, Chen, Guo, Zhang, Wang, and Wang}]{zhao2024we}
Zhonghan Zhao, Ke~Ma, Wenhao Chai, Xuan Wang, Kewei Chen, Dongxu Guo, Yanting Zhang, Hongwei Wang, and Gaoang Wang. 2024.
\newblock Do we really need a complex agent system? distill embodied agent into a single model.
\newblock \emph{arXiv preprint arXiv:2404.04619}.

\bibitem[{Zhou et~al.(2024)Zhou, Xu, Zhu, Zhou, Lo, Sridhar, Cheng, Ou, Bisk, Fried, Alon, and Neubig}]{webarena}
Shuyan Zhou, Frank~F. Xu, Hao Zhu, Xuhui Zhou, Robert Lo, Abishek Sridhar, Xianyi Cheng, Tianyue Ou, Yonatan Bisk, Daniel Fried, Uri Alon, and Graham Neubig. 2024.
\newblock \href {https://openreview.net/forum?id=oKn9c6ytLx} {Webarena: {A} realistic web environment for building autonomous agents}.
\newblock In \emph{The Twelfth International Conference on Learning Representations, {ICLR} 2024, Vienna, Austria, May 7-11, 2024}. OpenReview.net.

\end{thebibliography}
\appendix
\label{sec:appendix}
\section{Details of WebCoT Implementation}
\label{sec:webcot_implementation}
We provide additional details about the prompts used for the web agent in our experiments. The prompts are categorized as follows: \textbf{\{Prompt Refinement Hints\}}, \textbf{\{Environment Description\}}, and \textbf{\{Agent Hints\}}. These are provided in Table~\ref{tab:prompt_refine_prompt}, Table~\ref{tab:env_description_prompt}, and Table~\ref{tab:agent_hints_prompt}, respectively. The system-level prompt used for web agent action generation is detailed in Table~\ref{tab:agent_sys_prompt}. For the automatic evaluation conducted using \textsc{GPT-4o}, the evaluation prompt is listed in Table~\ref{tab:eval_gpt_prompt}. 

For the \textit{Reflection \& Lookahead} mechanism, the prompt used to construct $h^L_t$ is provided in Table~\ref{tab:lookahead_cot_prompt}. Similarly, the prompt for the \textit{Rollback} mechanism, which involves constructing the rollback CoT, is shown in Table~\ref{tab:backtracking_prompt}. 

All OpenAPI calls were configured with the following parameter settings: \texttt{max\_tokens} was set to 1,000, the random seed was fixed at 42, and the temperature was set to 0 to ensure reproducibility. 

Regarding inference settings, \texttt{max\_tokens} was set to 10,240 for \textsc{QwQ-32B} and \textsc{Llama-3.3-70B + QwQ Distill} to accommodate their extended reasoning capabilities, while it was limited to 2,048 for all other model variants. Across all experiments, the temperature parameter was fixed at 0 to ensure deterministic outputs and reproducibility.

In Table~\ref{tab:token_and_pricing}, we show the number of tokens and API price induced for the whole WebCoT. This results in exceptionally low operational costs, making our approach scalable, cost-effective, and practical for real-world applications.
\begin{table}[!ht]
    \centering
    \setlength{\tabcolsep}{2mm}{
    \begin{tabular}{l|cc}
    \toprule
         Method  
         & \textsc{Token Num.} & \textsc{Price~\$} \\
         \midrule
         \textsc{Reflection} & 5,943,491 & 26.15\\
         \textsc{Branching}  & 289,886 & 1.28\\
         \textsc{Rollback}  & 54,233 & 0.24\\
         \textsc{Total} & 6,287,610 & 27.67\\
    \bottomrule

    \end{tabular}
    }
    \caption{Generated token number and price (in USD) from using \textsc{GPT-4o-mini} in WebCoT.}
    \vspace{-0.4cm}
    \label{tab:token_and_pricing}
\end{table}
\begin{table}[!h]
    \centering
    \small
    \renewcommand{\arraystretch}{1.1}
    \setlength{\tabcolsep}{1.6mm}{
    \begin{tabular}{p{2.3cm}ccc}
    \toprule
         \multirow{1}{*}{Method}
         & WebVoyager & M2W & SimQA \\
         \midrule
         \textsc{Rej. Sampling}  & {22.08} & {46.00} & {16.98} \\
         \textsc{GPT-4o Distill} & {22.59} & {22.00} & {16.98}\\
         \textsc{WebCoT} & \textbf{25.52} & \textbf{47.00} & \textbf{18.87}\\
         \bottomrule
    \end{tabular}
    }
    \caption{Performance of WebCoT on Qwen3 8B.}
    \vspace{-0.5cm}
    \label{tab:qw3_8b_perf}
\end{table}
\begin{table*}[!ht]
\small
    \centering
    \begin{tcolorbox}[colframe=black, colback=gray!10!white, coltitle=black, boxrule=0.5mm]
     $\mathrm{\mathbf{\{Environment\ Description\}}}$\\
     \\
     $\mathrm{\mathbf{\{Prompt\ Refinement\ Hints\}}}$
    \\\\
    Chain of Thought demonstration: $\{h_{t_c}\}$\\
    The task: $\{q\}$\\
    The navigation history: $\{\tau\}$\\
    The current observation (web page's accessibility tree): $\{\tau_{t<t_c}\}$\\
    The current action you are about to exactly choose: $\{a_{t_c}\}$\\
    The navigation lookahead: $\{\tau_{t>t_c}\}$\\
    \\
    Please directly generate your thoughts and critiques.

    \end{tcolorbox}
    \caption{Prompt for constructing $h^L_{t}$.}
    \label{tab:lookahead_cot_prompt}
\end{table*}

\begin{table*}[!ht]
\small
    \centering
    \begin{tcolorbox}[colframe=black, colback=gray!10!white, coltitle=black, boxrule=0.5mm]
    Please directly generate your chain of thoughts and critiques, and reasoning right before exactly choosing the given current action $\{a_{t_c}\}$ according to the task, the navigation history/lookahead, and the current observation. Your thoughts should be focused on: What important information for the task completion can be expected after performing the current action based on the current observation within the broader navigation context? How does the current action, based on the current observation, contribute to achieving the overall task goal within the broader context of the navigation overview? How necessary is the current action based on the current observation for the task completion in the context of the overall navigation overview? Additionally, provide a detailed plan outlining the next steps after completing the current action, ensuring it aligns with the navigation overview. \\\\
    \textbf{Hints:}\\
    1. Be aware of the task’s constraints while offering your insights. \\
    2. Try to avoid mentioning the current action at the beginning of the chain of thought.\\
    3. Write the chain of thought supposing that the given current action has not been taken, and you are giving a look-ahead of what will happen in the future.\\
    4. Your chain of thought should be shorter as length of navigation lookahead decreases, which means you are closer to the task completion.\\

    \end{tcolorbox}
    \caption{Prompt for $\mathrm{\mathbf{\{Prompt\ Refinement\ Hints\}}}$.}
    \label{tab:prompt_refine_prompt}
\end{table*}

\begin{table*}[!ht]
\small
    \centering
    \begin{tcolorbox}[colframe=black, colback=gray!10!white, coltitle=black, boxrule=0.5mm]
    You are an autonomous intelligent agent tasked with navigating a web browser. You will be given web-based tasks. These tasks will be accomplished through the use of specific actions you can issue.\\
    Here's the information you'll have:\\
    \textbf{The user's objective:} This is the task you're trying to complete.\\
    \textbf{The current observation (web page's accessibility tree):} This is a simplified representation of the webpage, providing key information. Optionally, you may be provided with a screenshot of the webpage. You should pay close attention to the screenshot to make decisions.\\
    \textbf{The open tabs:} These are the tabs you have open.\\
    \textbf{The previous actions:} You can refer to the conversation history with the user to see the actions you have taken. It may be helpful to track your progress.\\
    \\
    The actions you can perform are the following:\\
    \textbf{`click [id]`:} This action clicks on an element with a specific id on the webpage.\\
    \textbf{`type [id] [content] [press\_enter\_after=0|1]`:} Use this to type the content into the field with id. By default, the "Enter" key is pressed after typing unless press\_enter\_after is set to 0.\\
    \textbf{`scroll [direction=down|up]`:} Scroll the page up or down.\\
    \textbf{`goback`:} Navigate to the previously viewed page.\\
    \textbf{`restart`:} Navigate to the original homepage at first. When you can't find information on some websites, try starting over from the beginning.\\
    \textbf{`stop [answer]`:} Issue this action when you believe the task is complete. If the objective is to find a text-based answer, provide the answer in the bracket. If you believe the task is impossible to complete, provide the answer as "N/A" in the bracket.

    \end{tcolorbox}
    \caption{Prompt for $\mathrm{\mathbf{\{Environment\ Description\}}}$.}
    \label{tab:env_description_prompt}
\end{table*}

\begin{table*}[!ht]
\small
    \centering
    \begin{tcolorbox}[colframe=black, colback=gray!10!white, coltitle=black, boxrule=0.5mm]
    To be successful, it is very important to follow the following rules:\\
    1. If you are uncertain about the next action, follow these steps: First, generate up to three of the most likely and valid actions based on the current observation. Then, for each of these possible actions, simulate and describe the expected future outcome in free text, detailing the next observation that would result from performing the action. Next, evaluate the correctness of each action by considering both the current observation and the simulated future results. Assign a numerical score from 0 to 1 to indicate the likelihood of correctness for each action: a score of 1.0 means "complete", 0.5 means "on track", and 0 means "incorrect". Provide your rationale for each score before assigning it. Finally, select and output the action with the highest score from the evaluated actions.\\
    2. You should only issue an action that is valid given the current observation. For example, you should NOT type into buttons or click on statictext.\\
    3. You should only issue one action at a time.\\
    4. STRICTLY Avoid repeating the same action if the webpage remains unchanged. You may have selected the wrong web element or numerical label.\\
    5. Issue stop action when you think you have achieved the objective. Don't generate anything after stop.\\
    6. If you ever need to login, login with Google. Try to skip any follow-up questions that may appear after logging in.\\
    Your reply should strictly follow the format:\\
    \\
    \textbf{<think>}\\
    1. \textit{Thought:} \{\{Your brief thoughts (briefly summarize the info that will help complete the task)\}\}\\
    \textit{Possible Step:} \{\{One of the logical and valid actions to take based on the current observation.\}\}\\
    \textit{Simulated Output:} \{\{A prediction of what the next observation or result will be after performing the action.\}\}\\
    \textit{Critic Evaluation:} \{\{Your rationale on the effectiveness of the action as well as a score from 0 (poor performance) to 1 (excellent performance), judging the corresponding action's s effectiveness.\}\}\\
    2. ... (continue with subsequent steps as needed in the same format)\\
    \textbf{</think>} (Optional: You can choose to include the steps between `<think>` and `</think>` if necessary or skip them based on the task's complexity.)\\\\
    
    Thought: {{Your brief thoughts (briefly summarize the info that will help complete the task)}} Action: ```{{The final action you choose to take in the process.}}```\\

    \end{tcolorbox}
    \caption{Prompt for $\mathrm{\mathbf{\{Agent\ Hints\}}}$.}
    \label{tab:agent_hints_prompt}
\end{table*}

\begin{table*}[!ht]
\small
    \centering
    \begin{tcolorbox}[colframe=black, colback=gray!10!white, coltitle=black, boxrule=0.5mm]
    As an evaluator, you will be presented with three primary components to assist you in your role:\\
    1. Web Task Instruction: This is a clear and specific directive provided in natural language, detailing the online activity to be carried out. These requirements may include conducting searches, verifying information, comparing prices, checking availability, or any other action relevant to the specified web service (such as Amazon, Apple, ArXiv, BBC News, Booking etc).\\
    2. Result Webpage Accessibility Tree: This is a representation of the web page showing the result or intermediate state of performing a web task. It serves as proof of the actions taken in response to the instruction.\\
    3. Result Response: This is a textual response obtained after the execution of the web task. It serves as textual result in response to the instruction.\\
    \\
    -- You DO NOT NEED to interact with web pages or perform actions such as booking flights or conducting searches on websites.\\
    -- You SHOULD NOT make assumptions based on information not presented in the webpage when comparing it to the instructions.\\
    -- Your primary responsibility is to conduct a thorough assessment of the web task instruction against the outcome depicted in the screenshot and in the response, evaluating whether the actions taken align with the given instructions.\\
    -- NOTE that the instruction may involve more than one task, for example, locating the garage and summarizing the review. Failing to complete either task, such as not providing a summary, should be considered unsuccessful.\\
    -- NOTE that the screenshot is authentic, but the response provided by LLM is generated at the end of web browsing, and there may be discrepancies between the text and the screenshots.\\
    -- Note the difference:\\
    1) Result response may contradict the screenshot, then the content of the screenshot prevails, 2) The content in the Result response is not mentioned on the screenshot, choose to believe the content.\\
    \\
    You should elaborate on how you arrived at your final evaluation and then provide a definitive verdict on whether the task has been successfully accomplished, either as 'SUCCESS' or 'NOT SUCCESS'.
    \end{tcolorbox}
    \caption{Prompt for \textbf{GPT-4o} automatic evaluation.}
    \label{tab:eval_gpt_prompt}
\end{table*}
\begin{table*}[!ht]
\small
    \centering
    \begin{tcolorbox}[colframe=black, colback=gray!10!white, coltitle=black, boxrule=0.5mm]
     \textbf{<think>}\\
    \ \ \ ......\\
    \ \ \ $k$. Thought: $\{h^{(k)}_t\}$\\
    \ \ \ \ \ \ \ Possible Step: $\{a^{(k)}_t\}$\\
    \ \ \ \ \ \ \ Simulated Output: $\{sim(o_j, a^{(k)}_t)\}$\\
    \ \ \ \ \ \ \ Critic Evaluation: $\{score(sim(o_j, a^{(k)}_t))\}$\\
    \textbf{</think>}\\
    \\
    Thought: $\{h^{(I)}_t\}$\\

    \end{tcolorbox}
    \caption{Template of constructing $h^{B}_t$.}
    \label{tab:branch_prompt}
\end{table*}

\begin{table*}[!ht]
\small
    \centering
    
    \begin{tcolorbox}[colframe=black, colback=gray!10!white, coltitle=black, boxrule=0.5mm]
     $\mathrm{\mathbf{\{Environment\ Description\}}}$\\\\
     $\mathrm{\mathbf{\{Agent\ Hints\}}}$\\\\
     Previously, the action "$\{a_j\}$" has been attempted. Please explore a different action.
    \end{tcolorbox}
    \caption{Prompt for generating alternative thoughts and actions.}
    \label{tab:alternative_actions_prompt}
\end{table*}

\begin{table*}[!ht]
\small
    \centering
    
    \begin{tcolorbox}[colframe=black, colback=gray!10!white, coltitle=black, boxrule=0.5mm]
     $\mathrm{\mathbf{\{Environment\ Description\}}}$\\\\
     $\mathrm{\mathbf{\{Agent\ Hints\}}}$
    \end{tcolorbox}
    \caption{System prompt for web agent.}
    \label{tab:agent_sys_prompt}
\end{table*}

\begin{table*}[!ht]
\small
    \centering
    
    \begin{tcolorbox}[colframe=black, colback=gray!10!white, coltitle=black, boxrule=0.5mm]
     $\mathrm{\mathbf{\{Environment\ Description\}}}$\\\\
     $\mathrm{\mathbf{\{Agent\ Hints\}}}$\\\\
     Previously, the action "$\{a'_j\}$" has been attempted, and this action will not lead to the task completion. Please provide an action for going back to the last observation following the aforementioned format. Give your brief reason why this action cannot help to complete the task.\\\\
     Last Observation: $\{o_j\}$\\
     Current Observation: $\{o'_{j+1}\}$
    \end{tcolorbox}
    \caption{Prompt for constructing Rollback CoT.}
    \label{tab:backtracking_prompt}
\end{table*}

\section{Implementation Details of WebDreamer}
\label{sec:webdreamer}
The prompt designs for proposing possible actions, ${(h^{(i)}_t, a_t^{(i)})}^k_{i=1}$, simulating the outcomes of these actions, and evaluating them based on the simulations are provided in Table~\ref{tab:act_prop_wd_prompt}, Table~\ref{tab:wd_simluation_prompt}, and Table~\ref{tab:wd_eval_prompt}, respectively.
\begin{table*}[!ht]
    \small
    \centering
    \begin{tcolorbox}[colframe=black, colback=gray!10!white, coltitle=black, boxrule=0.5mm]
     $\mathrm{\mathbf{\{Environment\ Description\}}}$\\\\
     $\mathrm{\mathbf{\{Agent\ Hints\}}}$\\\\
     Please generate actions different from $\{(h_t^{(i)}, a_t^{(i)})\}^{k-1}_{i=1}$.
    \end{tcolorbox}
    \caption{Prompt for proposing $(h_t^{(k)}, a_t^{(k)})$ in WebDreamer.}
    \label{tab:act_prop_wd_prompt}
\end{table*}
\begin{table*}[!ht]
    \small
    \centering
    \begin{tcolorbox}[colframe=black, colback=gray!10!white, coltitle=black, boxrule=0.5mm]
    You are a web server. You are given the current observed accessibility tree of the web page, and an action to perform. The expected output is a short description on what the next observation is, in the form of free text.\\
    \\
    The definitions of the actions are as follows: The actions you can perform are the following:\\
    \textbf{`click [id]`:} This action clicks on an element with a specific id on the webpage.\\
    \textbf{`type [id] [content] [press\_enter\_after=0|1]`:} Use this to type the content into the field with id. By default, the "Enter" key is pressed after typing unless press\_enter\_after is set to 0.\\
    \textbf{`scroll [direction=down|up]`:} Scroll the page up or down.\\
    \textbf{`goback`:} Navigate to the previously viewed page.\\
    \textbf{`restart`:} Navigate to the original home page and restart the action.
    \end{tcolorbox}
    \caption{Prompt for action simulation in WebDreamer.}
    \label{tab:wd_simluation_prompt}
\end{table*}
\begin{table*}[!ht]
    \small
    \centering
    \begin{tcolorbox}[colframe=black, colback=gray!10!white, coltitle=black, boxrule=0.5mm]
     You are an evaluator of a web agent task, evaluating the correctness of the action, conditioned on the current observation and a simulated future result.\\
     You are given the task query, the current observed accessibility tree, the action performed, and a textual description of the simulated output after performing this action.\\
     You are expected to give a numerical score (0 to 1) to indicate whether the simulated output is correct. The higher the score, the more likely the action is correct.\\
     \\
     Here are some example scores: complete (1.0), on track (0.5), or incorrect (0).\\
     Output your rationale first and then the score.\\
     \\
     Output format:\\
     Thought: XXXX. Score: \{a score from 0 to 1\}.
    \end{tcolorbox}
    \caption{Prompt for action evaluation in WebDreamer.}
    \vspace{-0.5cm}
    \label{tab:wd_eval_prompt}
\end{table*}

\section{Details of Finetuning}
We use the Megatron-LM\footnote{https://github.com/NVIDIA/Megatron-LM} framework to finetune all the models. On each dataset, one epoch is trained, using a learning rate of 1e-5 and a batch size of 64.

\section{Performance of WebCoT on Qwen3-8B}
To assess WebCoT's broader applicability, we conducted supplementary experiments using \textsc{Qwen3-8B-Instruct}. The results in Table~\ref{tab:qw3_8b_perf} show that WebCoT maintains its performance advantage over rejection sampling across all evaluated benchmarks. These findings demonstrate that our fine-tuning approach generalizes beyond \textsc{LlaMA-3.3-70B-Instruct}, confirming WebCoT's robustness.

\section{Performance of GPT-4o Distillation}
We also present the results of the baseline performance of \textsc{Qwen3-8B} distilled from \textsc{GPT-4o} trajectories in Table~\ref{tab:qw3_8b_perf}. Specifically, we sampled successful trajectories generated by GPT-4o on the training queries of the WebVoyager dataset and fine-tuned the Qwen3 model under identical experimental settings.
The results demonstrate that WebCoT offers a clear advantage over GPT-4o distillation on the WebVoyager dataset. Interestingly, we observe that GPT-4o distillation harms OOD performance in the SimpleQA setting compared to other baselines. This finding reinforces the conclusion that WebCoT fundamentally enhances the generalizability of LLMs by introducing critical cognitive behaviors absent in other approaches.

\section{Case Study of Hallucination}
Table~\ref{tab:hallu_case} presents two common types of hallucination errors observed in the web agent fine-tuned on $\hat{\mathcal{D}}^c_L$. In the first example, the agent fabricates the statement: ``the price starts at \$799 for the 128G model,'' despite the fact that ``128G'' does not exist in the webpage observation. The second example demonstrates a hallucination in the accessibility tree, where the agent incorrectly assumes the existence of an element with ID [11] to click, whereas the webpage's maximum element ID is only 5.
\label{sec:hallu_case_study}
\begin{table*}[!ht]
\small
    \centering
    
    \ttfamily
    \begin{tcolorbox}[colframe=black, colback=gray!10!white, coltitle=black, boxrule=0.5mm]
     \textit{OBJECTIVE:} latest iPhone model price and screen size comparison between iPhone Pro and iPhone Pro Max\\
     \textit{OBSERVATION:} Tab 0 (current): Apple[1] RootWebArea 'iPhone - Apple' focused: true[2] navigation 'Global'[3] link 'Apple'[4] link 'Store'[5] button 'Store menu' expanded: false[6] link 'Mac'[7] button 'Mac menu' expanded: false[8] link 'iPad'[9] button 'iPad menu' expanded: false[10] link 'iPhone'[11] button 'iPhone menu' expanded: false[12] link 'Watch'[13] button 'Watch menu' expanded: false[14] link 'Vision'[15] button 'Vision menu' expanded: false[16] link 'AirPods'[17] button 'AirPods menu' expanded: false[18] link 'TV and Home'[19] StaticText 'TV \& Home'[20] button 'TV and Home menu' expanded: false[21] link 'Entertainment'[22] button 'Entertainment menu' expanded: false[23] link 'Accessories'[24] button 'Accessories menu' expanded: false[25] link 'Support'[26] button 'Support menu' expanded: false[27] button 'Search apple.com'[28] button 'Shopping Bag'[29] link 'iPhone 16 Pro'[30] link 'iPhone 16'[31] link 'iPhone 16e New'[32] link 'iPhone 15'[33] link 'Compare'[34] link 'AirPods'[35] link 'AirTag'[36] link 'Accessories'[37] link 'Apple Card'[38] link 'iOS 18'[39] link 'Shop iPhone'[40] StaticText 'Get \$170–'[41] StaticText '\$630 in credit toward iPhone 16 or iPhone 16 Pro when you trade in iPhone 12 or higher.'[42] link 'Footnote * symbol'[43] link 'Shop iPhone'[44] StaticText 'Designed to be loved.'[45] button 'Play iPhone 16 lineup animation'\\
     \textit{THOUGHT:} The current observation indicates that I am on the page for the iPhone 16 Pro, which is the latest model in the iPhone series. The page provides essential details about the device, including its price, which is listed as starting at \$799 for the  \textcolor{red!70!black}{128GB} model. This information is crucial for completing the task, as it directly answers the question regarding the latest model and its price. Given that the task is to find the latest model of the iPhone and its price, the current action of stopping and providing the answer is both necessary and sufficient. The navigation history shows that I have successfully navigated to the relevant page and retrieved the required information. There are no further actions needed, as I have already obtained the answer to the task. After completing this action, there are no additional steps required, as the task will be considered complete. The answer will be provided in the specified format, and I can conclude the navigation process. In summary, the current action is essential for task completion, and since I have all the necessary information, I can confidently proceed to stop and provide the answer.\\
     \textit{ACTION:} ```stop [The latest model of the iPhone is the iPhone 16 Pro, and its price starts at \$799 for the  \textcolor{red!70!black}{128GB} model.]```
    \tcbline
    \textit{OBJECTIVE:} A 175cm tall, 85kg, 40yo man climbs 2500 steps at about 18cm per step and 40 steps per minute metabolic properties"\\
    \textit{OBSERVATION:} Tab 0 (current): Wolfram|Alpha: Computational Intelligence[1] RootWebArea 'Wolfram|Alpha: Computational Intelligence' focused: true[2] button 'UPGRADE TO PRO' hasPopup: menu expanded: false[3] button 'APPS' hasPopup: menu expanded: false[4] link 'TOUR'[5] button 'Sign in'\\
    \textit{THOUGHT:} The current observation indicates that the input query regarding the energy expenditure of a 40-year-old male has been successfully entered into the Wolfram Alpha input field. \textcolor{red!70!black}{The next logical step is to click the "Compute input button" (identified as button [11])}, which will trigger the computation necessary to provide the answer to the query. By clicking this button, I expect to be directed to a results page that will contain the calculated energy expenditure for the specified activity. This action is crucial for progressing towards the task completion, as it directly leads to the acquisition of the desired information. Without this step, the task would remain unfulfilled, as I would not receive the answer needed.After performing the click action, the immediate next steps would involve examining the results presented on the following page. I would look for the specific answer regarding the energy expenditure, which should be prominently displayed. If the answer is clear and satisfactory, I would then proceed to stop the task and provide the final answer. If the results are unclear or do not address the question adequately, I may need to consider alternative actions, such as refining the query or seeking additional information. However, given the context, it is likely that the results will provide the necessary information to complete the task.\\
    \textit{ACTION:} \textcolor{red!70!black}{```click [11]```}
    \end{tcolorbox}
    \caption{Examples of hallucination in $\hat{\mathcal{D}}^c_L$. Parts with hallucination are marked in \textcolor{red!70!black}{red}.}
    \vspace{-0.5cm}
    \label{tab:hallu_case}
\end{table*}

\section{Case Comparison between WebCoT and Rejection Sampling}
Here we include a case comparison in Table~\ref{tab:case_study_webcot} demonstrating scenarios where WebCoT-optimized LLMs outperform baseline models in web navigation tasks. For example, in the task \textit{Find out which musician made the headlines in Music News}, the WebCoT variant exhibited a deeper understanding of the webpage interface. When the \textbf{"Music"} button was not explicitly available on the page, the WebCoT model correctly inferred that the relevant link could be found under the \textbf{"Culture"} section. In contrast, the rejection sampling baseline began to hallucinate and erroneously selected the \textbf{[15] "Travel"} link instead.
\begin{table*}[!ht]
    \small
    \centering
    \ttfamily
    \begin{tcolorbox}[colframe=black, colback=gray!10!white, coltitle=black, boxrule=0.5mm]
     \textbf{OBSERVATION:}\\
     Tab 0 (current): BBC News - Breaking news, video and the latest top stories from the U.S. and around the world\\
     \\
     $[1]$ RootWebArea 'BBC News - Breaking news, video and the latest top stories from the U.S. and around the world' focused: true\\ 
     - $[2]$ button 'Open menu'\\
     ......... \\
     - $[11]$ link 'Business'\\
     - $[12]$ link 'Innovation'\\
     - $[13]$ link 'Culture'\\
     - $[14]$ link 'Arts'\\
     \textcolor{red!70!black}{- $[15]$ link 'Travel'}\\
     - $[16]$ link 'Earth'\\
     - $[17]$ link 'Audio'\\
     - $[18]$ link 'Video'\\
     - $[19]$ link 'Live'\\
     ......... \\
     - $[31]$ link 'BBC InDepth'\\
     - $[32]$ button 'More'\\

     \textbf{OBJECTIVE:}
     Find out which musician made the headlines in Music News. Please interact with the BBC News website.\\
     \textbf{THOUGHT:}
     The current page is the homepage of BBC News, and the objective is to locate musician-related headlines in Music News. \textcolor{red!70!black}{To achieve this, I need to navigate to the "Music" section.}\\
     \textbf{ACTION:}
     \textcolor{red!70!black}{click [15]} \textit{(Incorrectly selects the "Travel" link due to hallucination.)}
    \tcbline
    \textbf{OBSERVATION:}\\
     Tab 0 (current): BBC News - Breaking news, video and the latest top stories from the U.S. and around the world\\
     \\
     $[1]$ RootWebArea 'BBC News - Breaking news, video and the latest top stories from the U.S. and around the world' focused: true\\ 
     - $[2]$ button 'Open menu'\\
     ......... \\
     - $[11]$ link 'Business'\\
     - $[12]$ link 'Innovation'\\
     \textcolor{cyan!70!blue}{- $[13]$ link 'Culture'}\\
     - $[14]$ link 'Arts'\\
     - $[15]$ link 'Travel'\\
     - $[16]$ link 'Earth'\\
     - $[17]$ link 'Audio'\\
     - $[18]$ link 'Video'\\
     - $[19]$ link 'Live'\\
     ......... \\
     - $[31]$ link 'BBC InDepth'\\
     - $[32]$ button 'More'\\

     \textbf{OBJECTIVE:}
     Find out which musician made the headlines in Music News. Please interact with the BBC News Website.\\
     \textbf{THOUGHT:}
     To find Music News, I need to navigate to the "Music" section. \textcolor{cyan!70!blue}{Although there is no direct link to "Music," the "Culture" link might lead to music-related news.}\\
     \textbf{ACTION:}
     \textcolor{cyan!70!blue}{click [13]} \textit{(Correctly selects the "Culture" link, demonstrating inference capability.)}
    \end{tcolorbox}
    \caption{Case comparison between rejected sampling baseline~(upper half) and WebCoT~(lower half). Parts with hallucination are marked in \textcolor{red!70!black}{red}, and correct ones are colored in \textcolor{cyan!70!blue}{cyan}.}
    \label{tab:case_study_webcot}
\end{table*}

\section{Additional Details on Mind2web-live and WebVoyager Dataset} \label{sec:omitted_sites}
We conduct our evaluations using a subset of the testing portion of Mind2Web-Live\footnote{\url{https://huggingface.co/datasets/iMeanAI/Mind2Web-Live/blob/main/mind2web-live_test_20241024.json}} and WebVoyager\footnote{\url{https://github.com/MinorJerry/WebVoyager/blob/main/data/WebVoyager_data.jsonl}}. 
The list of websites that are excluded is in Table~\ref{tab:omitted_websites}.

\begin{table*}[!ht]
\small
    \centering
    
    \begin{tcolorbox}[colback=red!5!white, colframe=red!75!black, breakable]
    \ttfamily
    \small
    EXCLUDED\_WEBSITES\_MIND2WEB = \{
    'exploretock', 'kohls', 'united', 'parking', 'viator', 'delta', 'redbox', 'soundcloud', 'gamestop', 'travelzoo', 'amctheatres', 'ryanair', 'cargurus', 'resy', 'rentalcars', 'kbb', 'cabelas', 'menards', 'yellowpages', 'tripadvisor', 'tiktok.music', 'stubhub', 'thumbtack', 'weather', 'uhaul', 'health.usnews', 'healthgrades', 'theweathernetwork', 'zocdoc', 'usnews.education', 'epicurious', 'osu.edu', 'ups', 'dmv.virginia.gov', 'extraspace', 'finance.yahoo', 'pinterest',
    'sixflags', 'spothero', 'justice.gov', 'foxsports', 'ign', 'koa', 'tvguide', 'webmd', 'sports.yahoo', 'babycenter', 'tesla',
    \}

    EXCLUDED\_WEBSITES\_WEBVOYAGER = \{
    'booking', 'espn', 'amazon', 'google', 'googleflight', 'allrecipes', 'cambridgedictionary'
    \}
\end{tcolorbox}
    
    \caption{List of omitted websites.}
    \vspace{-0.5cm}
\label{tab:omitted_websites}
\end{table*}

\end{document}